\documentclass[letterpaper, 10pt, conference]{IEEEconf}  

\IEEEoverridecommandlockouts                              

\usepackage{algpseudocode}
\usepackage{algorithm}
\usepackage{graphicx}
\usepackage{dblfloatfix}
\graphicspath{{.}, {./figs/}, {./images/}}
\usepackage{subcaption} 
\usepackage{amsmath}
\let\proof\relax \let\endproof\relax
\usepackage{amsthm}
\usepackage{amssymb}
\usepackage{amsfonts}
\usepackage{dsfont}
\usepackage{bm}
\usepackage{mathrsfs}
\usepackage{mathtools}
\usepackage{xspace}
\usepackage{tabularx}
\usepackage{siunitx} 
\usepackage[usenames,svgnames,dvipsnames,table]{xcolor}
\usepackage{tikz}
\usepackage[h]{esvect} 
\usepackage{pgfplots}
\usepackage{svg}
\usetikzlibrary{tikzmark}
\pgfplotsset{compat=1.3}
\definecolor{series_runningtime}{rgb}{0.55,0.05,0.05}   
\definecolor{series_precisioninbits}{rgb}{0.05,0.05,0.55}
\definecolor{series_trimmed}{rgb}{0.05,0.55,0.05}
\definecolor{series_ratio}{rgb}{0.55,0.55,0.55}

\usepackage{pgfplots}
\pgfplotsset{compat=newest}

\usepackage{sidecap}
\title{
Optimizing pre-scheduled, intermittently-observed MDPs
}

\newcommand{\blockcomment}[1]{}
\newcommand{\gobble}[1]{}
\newcommand{\gobblexor}[2]{#2}

\newcommand{\commentColor}{SteelBlue}
\newcommand{\filteringColor}{BrickRed}
\newcommand{\graphHalfColWidth}{0.98\linewidth/2}
\newcommand{\xAcross}{0.99\linewidth/10}

\newcommand{\vartype}[1]{\ensuremath{\rm{#1}}}
\newcommand{\argtype}[1]{\ensuremath{\rm{\textbf{#1}}}}
\newcommand{\classtype}[1]{\ensuremath{\rm{\textbf{#1}}}}
\newcommand{\paramtype}[1]{\ensuremath{\rm{#1}}}
\newcommand{\functype}[1]{\textsc{#1}}

\newcommand{\strideBound}{\ensuremath{\sigma}} 
\providecommand{\dt}{\verythinspace t}

\newcommand{\comms}{\scalebox{0.6}{\ensuremath{\mathrm{ck}}}}
\newcommand{\exec}{\scalebox{0.6}{\ensuremath{\mathrm{ex}}}}
\newcommand{\alphalabel}{\ensuremath{\alpha}}
\newcommand{\myalpha}{\ensuremath{\alpha}}
\newcommand{\gamC}{\ensuremath{\gamma_{\exec}}\xspace}
\newcommand{\costC}{\ensuremath{C_{\exec}}\xspace}
\newcommand{\gamK}{\ensuremath{\gamma_{\comms}}\xspace}
\newcommand{\vcost}{\ensuremath{\vv{C}}\xspace}
\newcommand{\vcostK}{\ensuremath{\vcost_{\comms}}\xspace}
\newcommand{\vcostA}{\ensuremath{\vcost_{\alpha}}\xspace}
\newcommand{\vcostC}{\ensuremath{\vcost_{\exec}^{\gamma}}\xspace}
\newcommand{\vT}{\ensuremath{\vv{T}}}
\newcommand{\valueV}[5]{\ensuremath{V_{\scalebox{0.7}{${#1},\!{#5}$}}({#2},{#3}\,;{#4})}}
\newcommand{\valueVraw}[2]{\ensuremath{V_{\scalebox{0.7}{${#1},\!{#2}$}}}}
\newcommand{\valueVP}[5]{\ensuremath{V'_{\scalebox{0.7}{${#1},\!{#5}$}}({#2},{#3}\,;{#4})}}
\newcommand{\valueVPraw}[2]{\ensuremath{V'_{\scalebox{0.7}{${#1},\!{#2}$}}}}
\newcommand{\valueQ}[5]{\ensuremath{Q_{\scalebox{0.7}{${#1},\!{#5}$}}({#2},{#3}\,;{#4})}}
\newcommand{\valueQP}[5]{\ensuremath{Q'_{\scalebox{0.7}{${#1},\!{#5}$}}({#2},{#3}\,;{#4})}}
\newcommand{\valueQraw}[2]{\ensuremath{Q_{\scalebox{0.7}{${#1},\!{#2}$}}}}
\newcommand{\valueQPraw}[2]{\ensuremath{Q'_{\scalebox{0.7}{${#1},\!{#2}$}}}}
\newcommand{\vPi}{\ensuremath{\textcolor{black}{\vv{\pi}}\xspace}}
\newcommand{\vPiP}{\ensuremath{\textcolor{black}{\vv{\pi}'}\xspace}}
\newcommand{\optPi}{\ensuremath{\textcolor{black}{\vv{\pi}^\star\!}}\xspace}
\newcommand{\costOptPi}[1]{\ensuremath{\textcolor{black}{\protect\vv{\pi}_{#1}^\star\!}\,}\xspace}
\newcommand{\costOptPiRaw}[1]{\ensuremath{\protect\vv{\pi}_{#1}^\star\!}\,\xspace}
\newcommand{\knownPi}[1]{\ensuremath{\textcolor{black}{\protect\vv{\pi}_{\!#1}}}\xspace}
\newcommand{\knowns}{\ensuremath{K}\xspace}

\newcommand{\executionCostAlone}{\ensuremath{\mathbf{E}}\xspace} 
\newcommand{\checkinCostAlone}{\ensuremath{\mathbf{C}}\xspace} 
\newcommand{\executionCost}[2]{\ensuremath{\executionCostAlone({#1},{#2})}} 
\newcommand{\checkinCost}[2]{\ensuremath{\checkinCostAlone({#1},{#2})}} 

\newcommand{\plotAxisWidth}{1.55\linewidth}
\newcommand{\thirdAxisShift}{1.4cm}

\newcommand{\plotscale}{0.56}

\newcommand{\runningTimePlot}[1]{\runningTimePlotQC{#1}{99.96}}
\newcommand{\runningTimePlotQC}[2]{
    \begin{tikzpicture}[scale = \plotscale, inner sep=0.1em]
        \begin{axis}[
            height = 2.5in,
            width = \plotAxisWidth,
            xlabel={\textbf{Margin}},
            ylabel={\textcolor{series_runningtime}{\small\textbf{Time (seconds)}}},
            axis y line*=left,
            ymin=4,
            ymax=65,
            xtick=data,
            xmin=-0.005,
            scaled y ticks = false,
            x tick label style={scaled ticks=base 10:2}
            ]
            \addplot[color=series_runningtime, dashed, thick, mark=diamond*, mark options={color = series_runningtime, solid, thin}] table [y=time, col sep=comma] {#1};\label{plot_1_y1}
        \end{axis}

        \begin{axis}[
            height = 2.5in,
            width = \plotAxisWidth,
            axis y line*=right,
            axis x line=none,
            ymin=#2,ymax=100.001,
            ylabel={\textcolor{series_precisioninbits}{\small\textbf{Quality}}},
            ylabel style={rotate=180},
        ]
            \addplot[color=series_precisioninbits, solid, thick, mark=x, mark options={color = series_precisioninbits, solid, ultra thick}] table [y=quality, col sep=comma] {#1};\label{plot_1_y2}

        \end{axis}

        \pgfplotsset{every outer y axis line/.style={xshift=\thirdAxisShift}, every tick/.style={xshift=\thirdAxisShift}, every y tick label/.style={xshift=\thirdAxisShift} }
        \begin{axis}[
            height = 2.5in,
            width = \plotAxisWidth,
            legend style={font=\footnotesize},
            legend pos=south west,
            axis y line*=right,
            axis x line=none,
            ymin=0,ymax=103,
            ylabel={\textcolor{series_trimmed}{\small\textbf{Trimmed (\%)}}},
            ylabel style={rotate=180},
        ]
            \addlegendimage{/pgfplots/refstyle=plot_1_y1}\addlegendentry{Planning time}
            \addlegendimage{/pgfplots/refstyle=plot_1_y2}\addlegendentry{Pareto front quality}
            \addlegendimage{/pgfplots/refstyle=plot_1_y3}\addlegendentry{Proportion trimmed}
            
            \addplot[color=series_trimmed, dashed, thick, mark=x, mark options={color = series_trimmed, solid, ultra thick}] table [y=trimmed, col sep=comma] {#1};\label{plot_1_y3}

        \end{axis}
    \end{tikzpicture}
}

\newcommand{\runningTimePlotLen}[1]{
    \begin{tikzpicture}[scale = \plotscale, inner sep=0.1em]
        \begin{axis}[
            height = 2.5in,
            width = \plotAxisWidth,
            xlabel={\textbf{Schedule Length}},
            ylabel={\textcolor{series_runningtime}{\small\textbf{Normalized Time}}},
            axis y line*=left,
            ymin=-0,
            xtick=data,
            xmin=0.5,
            ]
            \addplot[color=series_runningtime, solid, thick, mark=diamond*, mark options={color = series_runningtime, solid, thin}] table [y=frac, col sep=comma] {#1};\label{plot_4_y1}
            \addplot[color=series_runningtime, dashed, thick, mark=diamond*, mark options={color = series_runningtime, solid, thin}] table [y=fracalpha, col sep=comma] {#1};\label{plot_4_y1_2}
        \end{axis}

        \begin{axis}[
            height = 2.5in,
            width = \plotAxisWidth,
            axis y line*=right,
            axis x line=none,
            ymin=60,ymax=101,
            ylabel={\textcolor{series_precisioninbits}{\small\textbf{Quality}}},
            ylabel style={rotate=180},
        ]
            \addplot[color=series_precisioninbits, solid, thick, mark=x, mark options={color = series_precisioninbits, solid, ultra thick}] table [y=quality, col sep=comma] {#1};\label{plot_4_y2}

        \end{axis}

        \pgfplotsset{every outer y axis line/.style={xshift=\thirdAxisShift}, every tick/.style={xshift=\thirdAxisShift}, every y tick label/.style={xshift=\thirdAxisShift} }
        \begin{axis}[
            height = 2.5in,
            width = \plotAxisWidth,
            legend style={at={(0.34,0.01)},anchor=south,font=\footnotesize},
            axis y line*=right,
            axis x line=none,
            ymin=-30,ymax=103,
            ylabel={\textcolor{series_trimmed}{\small\textbf{Trimmed (\%)}}},
            ylabel style={rotate=180},
        ]
            \addlegendimage{/pgfplots/refstyle=plot_4_y1}\addlegendentry{Normalized Time}
            \addlegendimage{/pgfplots/refstyle=plot_4_y2}\addlegendentry{Pareto front quality}
            \addlegendimage{/pgfplots/refstyle=plot_4_y3}\addlegendentry{Proportion trimmed}
            
            \addplot[color=series_trimmed, solid, thick, mark=x, mark options={color = series_trimmed, solid, ultra thick}] table [y=trimmed, col sep=comma] {#1};\label{plot_4_y3}
            \addplot[color=series_trimmed, dashed, thick, mark=x, mark options={color = series_trimmed, solid, ultra thick}] table [y=trimmedalpha, col sep=comma] {#1};\label{plot_4_y3_2}
            
        \end{axis}
    \end{tikzpicture}
}

\NewDocumentCommand{\LeftComment}{s m m}{%
\Statex \hspace{#2em} \textcolor{\IfBooleanTF{#1}{\filteringColor}{\commentColor}}{$\triangleright$} \parbox[t]{200pt}{\textcolor{\IfBooleanTF{#1}{\filteringColor}{\commentColor}}{#3}\strut} \hfill %
}

\algnewcommand\RightComment[1]{%
 \hfill \textcolor{\commentColor}{$\triangleright$ #1} %
}

\author{
Patrick Zhong$^{1}$, Federico Rossi$^{2}$, and Dylan A. Shell$^{1}$
\vspace*{-.3cm}
\thanks{$^{1}$Dept. of Comp. Sci. \& Eng., Texas A\&M University, College Station, TX 77843, USA.
        \scalebox{0.8}{\textsf{\{patrickzhong$\,|\,$dshell\}@tamu.edu}}}%
\thanks{$^{2}$Jet Propulsion Laboratory, California Institute of Technology, Pasadena, CA 91109, USA.
        \scalebox{0.8}{\textsf{federico.rossi@jpl.nasa.gov}}}%
\thanks{We acknowledge the support of Office of Naval Research Award \#N00014-22-1-2476. Part of this research was carried out at the Jet Propulsion Laboratory, California Institute of Technology, under a contract with the National Aeronautics and Space Administration (80NM0018D0004).}
}

\newtheorem{definition}{\textbf{Definition}}
\newtheorem{proposition}[definition]{\textbf{Proposition}}

\newenvironment{proofsketch}{%
  \proof}{\endproof}

\protected\def\verythinspace{%
  \ifmmode
    \mskip0.25\thinmuskip
  \else
    \ifhmode
      \kern .041670 em
    \fi
  \fi
}

\providecommand{\Reals}{\ensuremath{\mathbb{R}}}
\providecommand{\Nats}{\ensuremath{\mathbb{N}}}

\providecommand{\vs}{\textit{vs}\xspace}

\providecommand{\InitD}{\ensuremath{S_{0}}}

\providecommand{\observer}{observation process\xspace}
\providecommand{\actor}{actor\xspace}

\newlength\shlength
\newcommand\xshlongvec[2][0]{\ThisStyle{\setlength\shlength{#1\LMpt}%
  \stackengine{-5.6\LMpt}{$\SavedStyle#2$}{\smash{$\kern\shlength%
    \stackengine{\dimexpr 1.3pt+6.25\LMpt}{$\SavedStyle\mathchar"017E$}%
      {\rule{\widthof{$\SavedStyle#2$}}{\dimexpr.1pt+.5\LMpt}\kern.4\LMpt}{O}{r}{F}{F}{L}\kern-\shlength$}}%
      {O}{c}{F}{T}{S}}}

\newenvironment{defitems} {
    \begin{list}{$-$}{%
        \setlength{\leftmargin}{16pt}
        \setlength{\topsep}{-4pt}
        \setlength{\partopsep}{0pt}
        \setlength{\itemsep}{2pt}
        \setlength{\itemindent}{-6pt}}
        \ignorespaces}
{\unskip\end{list}}

\newif\ifmargincomments
\margincommentstrue

\ifmargincomments

\else

\fi

\begin{document}
\maketitle

\begin{abstract}
A challenging category of robotics problems arises when sensing incurs substantial costs.  
This paper examines settings in which a robot wishes to limit its observations of state, for instance, motivated by specific considerations of energy management, stealth, or implicit coordination. 
We formulate the problem of planning under uncertainty when the robot's observations are intermittent but their timing is known via a pre-declared schedule.
After having established the appropriate notion of an optimal policy for such settings, we tackle the problem of joint optimization of the cumulative execution cost and the number of state observations, both in expectation under discounts.
To approach this multi-objective optimization problem, we introduce an algorithm that can identify the Pareto front for a class of schedules that are advantageous in the discounted setting. 
The algorithm proceeds in an accumulative fashion, prepending additions to a working set of schedules and then computing incremental changes to the value functions.
Because full exhaustive construction becomes computationally prohibitive for moderate-sized problems, we propose a filtering approach to prune the working set.
Empirical results demonstrate that this filtering is effective at reducing computation while incurring only negligible reduction in quality. 
In summarizing our findings, we provide a characterization of the run-time \emph{vs} quality trade-off involved.
\end{abstract}

\section{Introduction}

We examine planning and control problems where obtaining a reliable
estimate of state can be costly or is generally undesirable. Unlike the existing body of work on
information gathering and active perception~\cite{bajcsy2018revisiting,aloimonos93active,bajcsy1988active}, the core question is not
\emph{what} or \emph{how} to sense, but rather \emph{when} to do so.  We
consider a setting in which the timing of observations must be pre-planned: the
robot is not merely permitted to decide online that it would be convenient to
receive a sensor reading now, but must pre-schedule the observations.
Far from being esoteric or abstruse, 
such problem instances
arise naturally: 

\subsubsection{Sensor network--enabled navigation}

Imagine a robot is navigating through a cave or subterranean cavern.  
Suppose that a mesh network of sensor motes is deployed within the space and
the robot queries the network to obtain its
pose by triangulating signal strengths (e.g., see~\cite{batalin2004mobile}).  
To boost longevity, the network nodes conserve energy by entering cycles of waking and hibernation.  
The robot can triangulate its position only when the nodes aren't
hibernating.  
Fortunately, one can specify, prior to deployment, the schedule by which nodes
set the watchdog timer to trigger their (synchronized) waking alarm.
Some schedules will be more informative than others; naturally, one wishes
to understand the relationship between the energy cost of the network and
navigational cost of the mobile robot.

\subsubsection{Stealth amidst snooping adversaries}

Consider an autonomous off-road vehicle moving in some GPS-denied
environment. The vehicle is supported by aircraft that fly overhead at
intervals, using their bird's-eye view to 
provide state information.
In the presence of nefarious entities wishing to harass either the
vehicle or the aircraft, it is important to 
find the right compromise between 
risk posed by more frequent flyovers \emph{vs} the precision
of off-road navigation.

\subsubsection{Relative rendezvous}

A pair of underwater gliders attempt to collect 
data in two parallel transects.  They resurface occasionally to
determine their poses relative to one another and to communicate, before diving
again to make additional measurements.
Treated as a single system with joint actions, 
what is an effective pre-determined ascent/descent
schedule, exploiting knowledge of the environment and their
task?


\subsection{Contributions and Paper Outline}

In the preceding, the robots all operate under uncertainty; they receive observations which are intermittent, but their sporadic occurrences have been scheduled beforehand.  
%
The three examples have the same core issue:
task performance generally improves with additional information but, though obtaining a state
estimate with high frequency diminishes uncertainty, the expense incurred 
(expressed as fuel/energy costs, or diminished stealth, or other factors)  
may outweigh those gains. 
%
The tension between these elements means that the question is 
how to find a suitable balance.


The perspective adopted in this paper is that appropriate compromises can be struck if informed by the Pareto frontier in the space of execution and observation costs.
Obtaining an exact representation this frontier is challenging and, hence,
we explore how to obtain an approximation in reasonable time.
In Sections~\ref{sec:formulation} and \ref{sec:model}, the paper formulates the class of problem, making both
notions of cost precise, as well as that of schedules. 
In Section~\ref{sec:algorithm}, 
we provide an algorithm that, starting from a collection of 
basic schedules, expands the set via a prepending operation along with an
incremental update to Bellman-like value functions.
The method includes several parameters which simplify its operation,
including a filtering approach that 
trades
resolution and 
completeness for running-time.
These aspects are explored through case studies and experiments in Sections~\ref{sec:character} and \ref{sec:results} of the paper.


\section{Related work}

Broadly speaking, choosing an observation schedule involves 
optimizing the perception process; this paper, thus,
represents an sort of perceptual optimization, realized via pre-planning.
We explore how choosing when to sense is important, 
and are constrained to have an ``open-loop declaration'' of this fact. 
This constraint makes our work rather distinct from
standard problems where most solutions use gathered information (or estimates)
to alter perception online. 

The present paper generalizes the work  presented in \cite{rossi22periodic}. In
that work, the authors assume that state observations are sparse, but will appear with a
strict periodicity: every $\kappa$ timesteps, for a specific and given $\kappa$.  
The algorithm in the present paper searches the space 
of schedules, a space containing many options 
besides the strictly periodic ones. And, as will be seen, the algorithm's
output can directly identify situations in which 
non-periodic schedules are superior.

The example scenarios earlier are all described as involving forms of
multi-agent interaction. 
For instance: the aircraft executing the overhead
flight generates an observation and communicates it with the ground
vehicle. The schedule forms a sort of communication 
timetable. Existing work, albeit with a quite different focus, 
examines the question of what and when to communicate in
planning settings, e.g.,~\cite{roth2006communicate,unhelkar2016contact}.

\section{Problem Formulation}
\label{sec:formulation}

\providecommand{\psmdp}{\textsc{ps-mdp}\xspace}
\providecommand{\psmdps}{\textsc{ps-mdp}s\xspace}
\providecommand{\psomdp}{\textsc{pso-mdp}\xspace}
\providecommand{\mdp}{\textsc{mdp}\xspace}
\providecommand{\mdps}{\textsc{mdp}s\xspace}
\providecommand{\sched}{\ensuremath{\mathcal{D}}\xspace}
\providecommand{\opt}{\ensuremath{\star}}
\providecommand{\optc}{\ensuremath{c}}
\providecommand{\idx}[1]{\ensuremath{\lceil{#1}\rfloor}}
\providecommand{\sS}{,} 
\providecommand{\seq}[1]{\ensuremath{({#1})}}
\providecommand{\recc}[1]{\ensuremath{\overline{#1}}}
\providecommand{\vect}[1]{\ensuremath{\bm{#1}}}
\providecommand{\allsched}{\ensuremath{\mathscr{D}}\xspace}

To begin, we define the class of pre-scheduled MDPs and
their solutions. We then give suitable costs associated with schedules,
formalize the core problem, and 
subsequently turn to some certain types of regularity on 
schedules.

\subsection{Preliminaries}

\begin{definition}[Schedules]
\label{def:sched}
A \emph{schedule} is a function $\sched: \Nats\cup\{0\} \to \Nats$ 
describing a sempiternal sequence\gobble{ of positive numbers}.
The schedule has \emph{stride bounded by $\strideBound$} if \mbox{$\sched(n) \leq \strideBound$}$, \forall n \in \Nats$.
Schedule $\sched$ is \emph{tail recurrent} if $\exists N_0 \in \Nats$\gobblexor{ such that }{, }\mbox{$\sched(n+1) = \sched(n)$}$, \forall n \geq N_0$. 
\end{definition}
We will \gobble{have occasion to }write schedules simply as sequences
of numbers, for example, consider 
$\sched = \seq{1\sS 3\sS 2\sS 1\sS 3\sS 2\sS 2\sS 2\sS 2\sS 2\sS \dots} \equiv \seq{1\sS 3\sS 2\sS 1\sS 3\sS \recc{2}}$.
To be compatible with the function definition, we will refer to elements 
with indexes starting from zero.
Clearly, the previous example given is tail recurrent.
To ease subsequent presentation, we use illustratory schedules with stride bounded by $9$, writing them concisely: $13213\recc{2}$.

\begin{definition}[\psmdp]
A \emph{pre-scheduled Markov Decision Process} 
is a tuple $\langle S, A, T, \costC, G, \sched \rangle$ where
\begin{defitems}
\item $S = \{s_1, s_2, \dots, s_{|S|}\}$ is the finite set of states; 
\item $A = \{a_1, a_2, \dots, a_{|A|}\}$ is the finite set of actions; 
\item $T: S\times A \times S \to [0,1]$ give the transition dynamics\gobble{, or
transition model,} describing the stochastic state transitions of the system,
assumed to be Markovian in the states, where: \qquad \phantom{......}$P(s^{t+1}\!=\!s'\,|\,s^{t}
= s, a^{t} = a) = T(s',a,s),\; \forall t,$
\item $\costC: S\times A\to \Reals$ is the function prescribing the cost incurred for taking action $a$ in state~$s$;
\item $G \subseteq S$ is a goal region;
\item $\sched$ is a schedule of state observations (or check-ins).\\[-4pt]
\end{defitems}
We will require that when $s \in G$,
for every $a \in A$,
 both $\costC(s,a) = 0$ and $T(s,a,s) = 1$.
\end{definition}

The first four elements are standard in the \mdp
literature~\cite{bertsekas19reinforcement,puterman2014markov,lavalle06planning}.
In classic instances, the optimal control problem involves, at each point in
time: (1) obtaining, via sensors, the robot's state, and (2) the decision of
which action to execute, and then (3) the robot undergoing a transition
satisfying the transition model.  
The \psmdp differs from the \mdp in the sparsity of
observations: states are disclosed to the robot, but only at specific points in
time. These times are those described by the schedule, $\sched$,
which provides an (infinite) sequence of natural numbers.  At time $t=0$, the
system obtains its state. Then $\sched(0)$ gives the number of time steps until
the state information will be obtained next; after that occurs, $\sched(1)$ is
the offset to the next observation, and so forth.  Thus, 
if $\sched = 
1111\dots = \seq{\recc{1}}$ then the \psmdp
is a classic \mdp, as, after each step, the state will be obtained in the very
next step.  When there is some $k$ so that $\forall n,  \sched(n) = k$, or
$\sched = \seq{k\sS k\sS k\sS \dots} = \seq{\recc{k}}$, then the \psmdp
corresponds to the periodically state-observed \mdp of
\cite{rossi22periodic}, with the check-in period equal to~$k$.

\subsection{Solutions to \psmdps}

Following the traditional performance measure for MDPs\gobble{ (and their kin)}, it is
natural to 
provide a discount factor $\gamC \in (0,1)$ and
consider the \emph{expected discounted cumulative cost} to reach a
goal state.  The expectation provides a concrete objective even though the process's
evolution is stochastic, allowing aspects of uncertainty, e.g., the initial
state may only be specified by a distribution $\InitD : S \to [0,1]$. 
Also, the 
dynamics expressed by $T$ are uncertain: given an action to perform,
the outcome is only known up to a distribution before being executed.
For standard \mdps, those actions come from 
a \emph{policy}, a map
$\pi:{S\to A}$, that gives an action for the realized state.
An optimal policy is one minimizing the performance measure.
The Markov property implies that the current state in an MDP suffices 
to determine the optimal action for each step.

A policy as a direct map from states to actions will not work for
\psmdps because, in general, the current state is not known 
at every timestep. 
Indeed, for schedule $\sched$,
states are disclosed to the robot only at times:
$\seq{0\sS  \sched(0)\sS  \sched(0)+\sched(1)\sS \dots\sS  \sum_{i=0}^{m-1} \sched(i)\sS  \dots}$.
To relate timestep with observation occurrences,
let $\idx{\dt}$ be the number of check-ins that have been received by the robot
by time $t$, that is
$\idx{\dt} \coloneqq \min \big\{ k:\Nats\;\vert\; \dt < \sum_{i=0}^{k-1}\sched(i)\big\}$.
A \psmdp policy (or just policy), is a 
\begin{equation}
\label{def:vpi}
\vPi : S \times \Nats \to A^\infty,
\text{ s. t. }
\vPi\big(s_{\dt}, \idx{\dt}\big) = \vv{a}^{\dt} \in A^{\sched(\idx{\dt}-1)},
\end{equation}
with
$A^\infty \coloneqq A \cup A^2 \cup A^3 \cup \cdots$, where
$A^2 \coloneqq A \times A$, 
$A^3 \coloneqq A \times A \times A$, and
$A^4 \coloneqq A \times A \times A \times A$, 
etc.
(The existence of such a policy for all \psmdps is proven in Section~\ref{sec:model}.)

For example, at $t=0$, if schedule {$\sched(\idx{0}-1)=\sched(0)=4$}, then 
$\vPi(s_{0},0)$ must yield an ordered sequence of $4$ actions.
Those actions will be executed by the robot, and thereafter 
it will obtain the next state, $s_{4}$. 
And $\idx{4}=2$, and the value $\sched(1)$ gives the 
delay (in units of time) until the state will be known next, so
$\vPi(s_{4},1)$ will give $\sched(1)$ actions.
Sequences of actions are generated as composite units, and executed
without state feedback between them. The dimension of these composites
is termed the \emph{stride}. Considered in order, the strides must match the
values in $\sched$.

\newpage
\section{The model}
\label{sec:model}

We consider a setting with two agents or parties: the \emph{\observer} and the \emph{\actor}.
Operationally, the \observer generates check-ins according to the
schedule while the \actor executes the policy. 
The \observer is interested in a notion of cost, distinct 
from $\costC$, but related to the number of check-ins required.
Note how $\sched$ forms part of the \psmdp's definition and, accordingly, the planning problem is 
solved \textsl{given} a schedule. It is sequential so, in game-like terms, the \observer selects
the schedule first and then declares it; the \actor seeks to minimize the cost $\costC$,
finding a policy subject to $\sched$.

For the \observer's cost,
consider the following:
\begin{equation}
\vcostK(s, \vv{a}) = \begin{cases} 
      0 & s \in G \\
      1 & s \notin G
   \end{cases}\textrm{ for all } s \in S, \vv{a} \in A^\infty.
\label{ref:cost}
\end{equation}

As $\vcostK : S \times A^\infty \to \Reals$, it considers sequences of actions---we
term such a function a \emph{macro cost}.
The unit penalty incurred in \eqref{ref:cost} 
models the fact that a check-in is generated at the end of a whole sequence of actions.
The longer the stride in a 
$\sched$, the more that penalty will be amortized across time steps (or, equivalently, elementary actions).

Though we have stated that the \actor's objective involves finding some cost minimizing $\vPi$, we must first
show that the concept in \eqref{def:vpi}
is indeed appropriate for \psmdps. 
This is point of Proposition~\ref{prop:opt-exists}.

\subsection{Policies for \psmdps}

First, we need the following two definitions:

\begin{definition}
\label{defn:t-composition}
For transitions $T: S\times A \times S \to [0,1]$, the
\emph{macro transition model} is the function $\vT: S\times A^\infty \times S \to [0,1]$ defined as

\vspace*{-3ex}
{
\begin{align*}
\vT\left(s',\vv{a}, s\right) &= 
\vT\left(s',(a_1, a_2, \dots, a_{|\vv{a}|}), s\right) \nonumber  \\
&= \!\!\!\!\!\!\!\!\!\sum_{\substack{(s_1, \dots, s_{|\vv{a}|})\in S^{|\vv{a}|}\\ \text{where } s_0 = s \text{ and } s_{|\vv{a}|} = s'}} \prod_{i=1}^{|\vv{a}|} T(s_{i+1},a_i,s_i).
\end{align*}
}
\end{definition}
\vspace*{-.3ex}

This is,  essentially, just convolving the basic transition dynamics; the
number of times is determined directly from the length of the $\vv{a}$ argument.
We can do the same thing to turn $\costC$ into a macro cost function:

\begin{definition}
\label{defn:costcompositions}
For discounting factor $\gamC \in (0,1)$ and cost
function $\costC: S\times A\to \Reals$, the corresponding \emph{macro execution cost} is
$\vcostC : S\times A^\infty \to\Reals$ defined as

\vspace*{-3ex}
{
\begin{align*}
&\vcostC\left(s,\vv{a}\right) = \vcostC\left(s,(a_1, a_2, \dots, a_{|\vv{a}|})\right)  \nonumber \\
&\qquad = \sum_{k=1}^{|\vv{a}|}\;\gamC^{(k-1)} \hspace*{-4.2ex} \sum_{\substack{(s_1, \dots, s_{|\vv{a}|})\in S^{|\vv{a}|}\\ \text{where } s_1 = s}} \hspace*{-4ex} \costC(s_k,a_k) \prod_{i=1}^{|\vv{a}|} T(s_{i+1},a_i,s_i). 
\end{align*}
}
\end{definition}
\vspace*{-.3ex}

Notice that the $\gamC$ ensures costs incurred later, owing to sequences of actions, are diminished
correctly.

Macro costs allow for the following notion of \psmdps policy evaluation.

\begin{definition}
\label{def:policy_eval}
The \emph{evaluation of policy}
$\vPi : S \times \Nats \to A^\infty$ 
from state $s$ 
on \psmdp $\mathcal{M}$ 
with respect to 
discount $\gamma \in (0,1)$ and macro cost
function $\vcost: S\times A^{\infty}\to \Reals$ is\\[-1.2ex]
\begin{align*}
\valueV{\vcost}{s}{\vPi}{\sched}{\gamma}\,  = & \;\;\vcost(s, \vPi(s)) \quad + \\
&\hspace{5ex} \gamma\!\!\sum_{s' \in S}{\vT(s', \vPi(s), s) \valueV{\vcost}{s'}{\vPi}{\sched}{\gamma}}.
\end{align*}
\end{definition}

In evaluating from some initial state distribution,
we will write 
$\valueV{\vcost}{\InitD}{\vPi}{\sched}{\gamma}$, for the expected cost with state $s^0 \sim \InitD$.
(The preceding has been defined for generic macro costs to be used when applied with 
\gamK, \vcostK 
and also 
\gamC, \vcostC.)
The following holds for macro costs in general too:

\begin{proposition}
\label{prop:opt-exists}
Given some \psmdp $\mathcal{M}$, 
discount $\gamma \in (0,1)$, macro cost
$\vcost: S\times A^{\infty}\to \Reals$, and
initial state distribution $\InitD$,
there always exists a 
stationary and deterministic \psmdp policy 
$\optPi : S \times \Nats \to A^\infty$ such that:\\[-4ex]

$$\valueV{\vcost}{\InitD}{\optPi}{\sched}{\gamma} = \inf\limits_{\vPi}\,\valueV{\vcost}{\InitD}{\vPi}{\sched}{\gamma}.$$
\end{proposition}
\begin{proofsketch}
Construct a countably infinite \mdp, associating a copy
of $S$ for each $\Nats \cup \{0\}$. 
Assuming the copies are indexed by $r$, the actions and costs are convolved
using the construction  
in Definition~\ref{defn:t-composition} and~\ref{defn:costcompositions},
restricted to actions with $|\vv{a}| = \sched(r)$, for each $r$.
Then, the existence of a traditional optimal policy $\pi^\opt$ on 
state-space $S\times \Nats$, producing composite actions with
stride $\sched(r)$ for states $(\,\cdot\,, r)$, is a suitable $\optPi$.
\end{proofsketch}

The preceding establishes that the formal objects involved do exist, 
have correct types and well-defined cost metrics.

\subsection{Problem: Schedule Dominance}\label{subsec:dominate}

The notational heaviness is because
we wish to express the evaluation of policies on different macro costs.
For some \psmdp $\mathcal{M} = \langle S, A, T, \costC, G, \sched \rangle$, 
execution discount \gamC, check-in discount \gamK, and
initial distribution $\InitD$, we say:
\begin{enumerate}
\item the \emph{execution cost} of the pair $(\sched, \vPi)$
is the expected \mbox{\gamC-discounted} cumulative cost to execute the policy:
    \[ \executionCost{\sched}{\vPi} = \valueV{\vcostC}{\InitD}{\vPi}{\sched}{\gamC};\]

\item the \emph{check-in cost} of the pair $(\sched, \vPi)$ 
is the expected \mbox{\gamK-discounted} number of check-ins until $G$ is attained:
    \[ \checkinCost{\sched}{\vPi} = \valueV{\vcostK}{\InitD}{\vPi}{\sched}{\gamK}.\]
\end{enumerate}
\smallskip

\label{ref:dominate} 

\noindent For a given initial state distribution \InitD, 
we will say some schedule $\sched_2$ \emph{dominates} $\sched_1$ if $\forall \vPi_1 \exists \vPi_2$ s.t. 
$\executionCost{\sched_2}{\vPi_2}  < \executionCost{\sched_1}{\vPi_1}$
and 
$\checkinCost{\sched_2}{\vPi_2}  < \checkinCost{\sched_1}{\vPi_1}$.
And also that $\sched_1$ is \emph{non-dominated} in some class of
schedules \allsched, if there is no other $\sched_2 \in \allsched$ that dominates $\sched_1$. 
To help improve practicality, Section~\ref{subsec:bounds} seeks to approximate this criterion via a set of scalarizations.

We can now state the general problem to solve:

\medskip
\noindent\textbf{General Problem.}
Let $S, A, T, \costC$, $G$, and $\InitD$ be given.
For a class of schedules \allsched, 
compute the non-dominated schedules on \psmdps, in terms of 
$\executionCost{\cdot}{\cdot}$ and~$\checkinCost{\cdot}{\cdot}$.

Note how this contrasts with standard treatment of a traditional MDP,
where the solution is only a policy, which
does not depend upon the initial
state distribution.


\subsection{Subclasses of Schedules}

If the problem is to be concrete, some specific and meaningful class \allsched of schedules must be identified;
for an implementation, 
some practical means is needed to describe
infinite schedules. It would be ideal, moreover, if 
(1.)~it was concise and convenient;
(2.)~howsoever the sequences are circumscribed, the restriction should `wash out' in the limit to infinity;
(3.)~the choice afforded opportunities to exploit problem-specific constraints.

Accordingly, as already introduced in Definition~\ref{def:sched}, we consider schedules with
two restrictions:

\emph{Bounded stride} can be useful to capture structure of particular problem
settings. For instance, in the rendezvous scenario, there is a maximum duration
before the underwater gliders must surface, which provides a bound naturally.  

\emph{Tail recurrence} holds for
any
schedule employing the bar notation solely on the last element,
for instance,
the earlier example $\sched = 
13213\recc{2}$ is tail recurrent.
Since the execution is discounted by $\gamma$, the effect of the tail diminishes quickly as prefixes are prepended.\footnote{Results presented subsequently will show this intuition to indeed be true.}

Now, using the preceding as a sort of `regularity condition' on schedules, we can state the focus of our algorithm.

\medskip
\noindent\textbf{Computational Problem.}
Let $S, A, T, \costC$, $G$, and $\InitD$ be given.
For the tail recurrent schedules with stride bounded by $\strideBound$,
compute the non-dominated schedules on \psmdps, in terms of 
$\executionCost{\cdot}{\cdot}$ and~$\checkinCost{\cdot}{\cdot}$.

\subsection{Bounds on Schedules' Costs} \label{subsec:bounds}

If it were easy to enumerate every possible policy, a given schedule's costs could be computed and a Pareto front for that schedule alone constructed; 
dominance across schedules could be determined by comparison of those fronts.
The question, thus, becomes one of determining,
in a computationally efficient way, the Pareto front
for each schedule.

\begin{SCfigure}
  \centering
  \caption{Example showing characterization of a schedule with policy $\protect\vPi$ as a point: \scalebox{0.95}{$(\executionCost{\sched}{\protect\vPi}, \checkinCost{\sched}{\protect\vPi})$}. Black lines mark the realizable fronts and red lines mark the optimistic fronts. 
  Solid fronts are the specific bounds with 
  $\costOptPi{\exec}$ and $\costOptPi{\comms}$,
  while dashed show the new fronts with the addition of an $\costOptPi{\alphalabel}$ 
  policy, with the uncertainty region shown in light blue.}
  \includegraphics[width=0.5\linewidth,trim={0 0 8cm 9cm},clip]{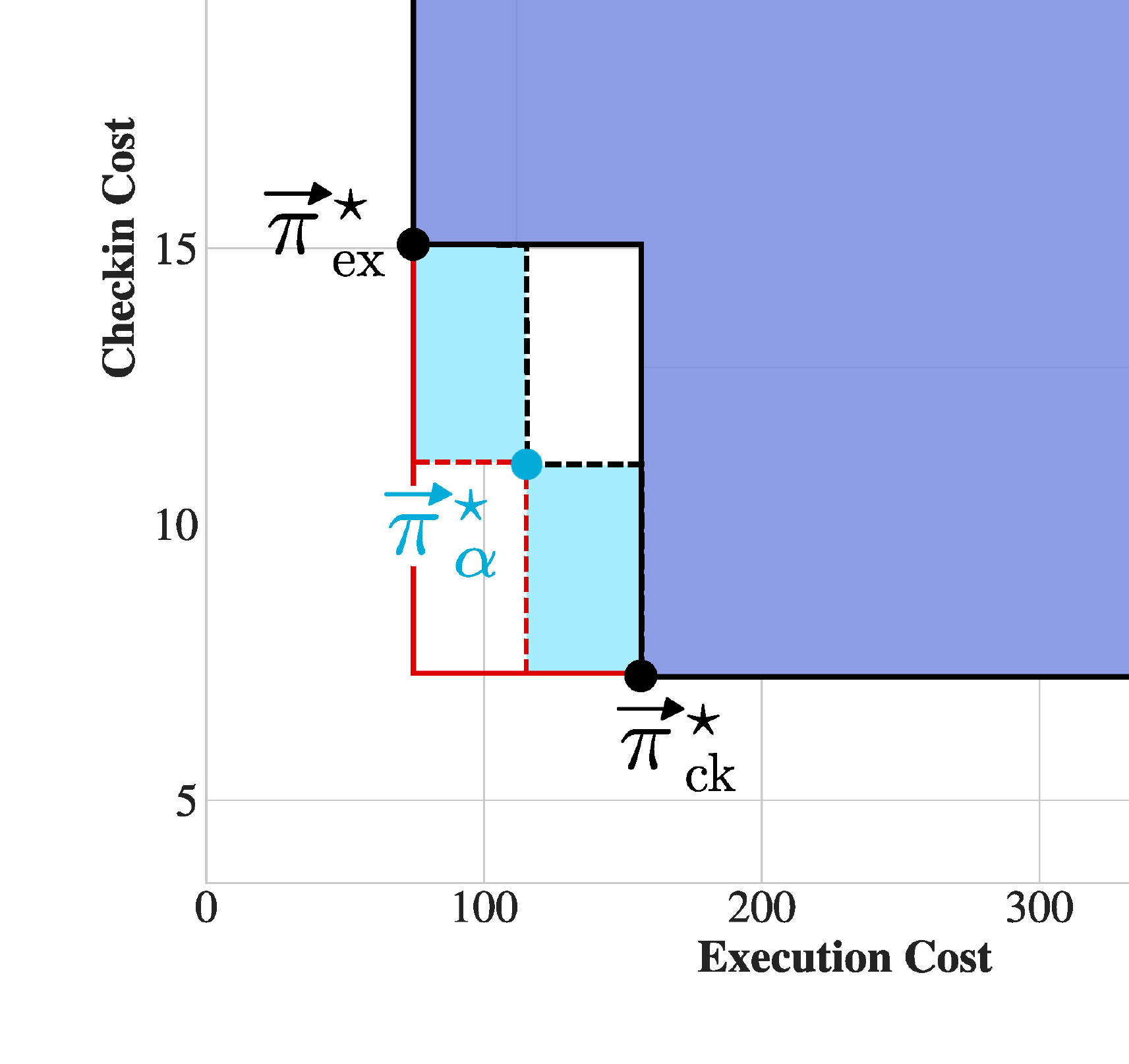}
  \label{fig:conceptual}
\vspace*{-2ex}
\end{SCfigure}

We can approximate the front of a schedule in the form of a bounded region
representing the best we \emph{know} a schedule can do and the best the
schedule can \emph{hypothetically} do. 
For a given schedule, there is a policy $\costOptPi{\exec}$ which minimizes the
execution cost \executionCostAlone, and a policy $\costOptPi{\comms}$ that minimizes the check-in cost \checkinCostAlone. We
can disregard any policy dominated by $\costOptPi{\exec}$ or $\costOptPi{\comms}$ ---\,a region
denoted in dark blue in Figure~\ref{fig:conceptual}. 
As 
$\costOptPi{\exec}$ 
is the
optimal policy with respect to execution cost, there can be no policies to its
left, and similarly there can be no policies below
$\costOptPi{\comms}$.
Of interest is the box formed with 
$\costOptPi{\exec}$ and $\costOptPi{\comms}$ 
as its corners:\gobble{, bounded from above and the right by a
\emph{realizable front} and bounded from below and the left by an
\emph{optimistic front}:}

\begin{definition}
For \psmdp 
$\langle S, A, T, \costC, G, \sched \rangle$ and a set of known policies $\knowns = \{\knownPi{1}, \knownPi{2}, \dots, \knownPi{|\knowns|}\}$, then
\begin{enumerate}
\item schedule \sched's \emph{realizable front} is the set of all policies in $\knowns$ not dominated by any other policy in $\knowns$;
\item schedule \sched's \emph{optimistic front} is the set of best possible (but unknown) policies such that none are dominated by any other possible policy.
\end{enumerate}

\end{definition}

One interpretation of the realizable front is an upper bound (in terms of cost) on the best case
scenario. One is able to guarantee the existence of a policy that dominates any
policies on or beyond the realizable front (and, when $\knowns = \{\costOptPi{\exec}\;,
\costOptPi{\comms}\,\}$, this is consistent  with what was expressed above).
The optimistic front corresponds to a lower bound on the best case scenario.
Unlike the realizable front, we have no guarantee that a policy on the
optimistic front will actually exist, but we do know that it is impossible to produce
any better policy.

The region bounded by these two fronts represents uncertainty in knowledge of the schedule's true front. We incorporate this by
considering a schedule $\sched_1$ dominated by another schedule $\sched_2$ if
each point in $\sched_1$'s optimistic front is dominated by some point in
$\sched_2$'s realizable front. The idea is that all optimal policies
in $\sched_1$, even hypothetical ones that may not actually exist, would be
worse than a realizable policy in $\sched_2$. 
This is safe treatment of dominance,
as no schedule is incorrectly marked as dominated, but
the price of uncertainty is in potentially
maintaining an excess of non-dominated schedules.

\subsection{Alpha Values}
Accurately comparing one schedule to another requires that we reduce the uncertainty region in each schedule. 
Computing every possible policy collapses the region down to the true front, but would be exceedingly expensive to compute.
Instead, we sample additional points in order to reduce (but not erase) the uncertainty. 
For a given $\myalpha \in (0, 1)$, we can produce a policy minimizing a blend of 
\executionCostAlone and \checkinCostAlone, representing a linear scalarization of the problem. 
The intuition is that, 
while $\costOptPi{\exec}$ 
minimizes along the horizontal and $\costOptPi{\comms}$ minimizes along the vertical,
the policy from an $\myalpha$-blend of costs will 
minimize along an intermediate line with the slope based on $\myalpha$; we denote the resulting policy by
$\costOptPi{\alphalabel}$.

The benefit of generating a new policy 
$\costOptPi{\alphalabel}$ is two-fold. 
Firstly, we have a new known policy that we can provide for the schedule (and hence include in \knowns), making it a new point in the realizable front. 
Secondly, since $\costOptPi{\alphalabel}$ is the optimal policy for its scalarization,
it is impossible for another policy to have both lower check-in and
lower execution cost, as that policy would then have a lower scalarized value.
The new policy, therefore, tightens the
lower bound by adding a point in the optimistic front. As
demonstrated by the blue $\costOptPi{\alphalabel}$ point and the dashed fronts in
Figure~\ref{fig:conceptual},  computing new policies this way decreases the
uncertainty region.

\section{Approach}
\label{sec:algorithm}

A na\"\i ve approach to computing the set of non-dominated schedules 
would be to enumerate all schedules,
computing each schedule's costs independently. 
Solving the problem in this way is prohibitively expensive due to exponential explosion. 
Hence, we next describe some insights to enable costs to be 
obtained much more efficiently.


\subsection{Schedule Substructure}
Firstly, while the number of schedules in the search space grows exponentially,
they are clearly not independent.  
A schedule $\seq{\sched(0),
\sched(1),\sched(2),\sched(3),\dots}$ can be divided into two parts, a prefix
$\sched(0)$ and a suffix $\seq{\sched(1),\sched(2),\sched(3),\dots}$. As the
suffix is itself also a schedule, we can define the costs of a schedule in
terms of the costs of its suffix. 
We can also consider a \psmdp policy $\vPi : S \times \Nats \to A^\infty$ as a series of policies $\pi_i : S \to A^{\sched(i)}$, with a policy for each stride in the schedule.
The Markov property implies that past policies in execution time depend on future policies but not vice versa, which we leverage by working backwards from the last policy to the first. Tail-recurrent schedules give us a natural starting point and base case---the recurrent tail $\recc{k}$, for which we can run standard value iteration (after employing
Definitions~\ref{defn:t-composition}--\ref{defn:costcompositions}) to
convergence for the last policy. We next show that the policy for the stride immediately before, representing $\sched = (\kappa, \recc{k})$,  can be incrementally computed through a process dubbed \emph{schedule extension}:


\begin{proposition}[Schedule extension]
Let \psmdp 
$\mathcal{M} = \langle S, A, T, \costC, G, \sched \rangle$ be given with
value function 
$\valueQraw{\vcost}{\gamma}$
and policy evaluation 
$\valueVraw{\vcost}{\gamma}$ for policy $\vPi$ under
discount $\gamma \in (0,1)$ and macro cost function $\vcost$.
Then value function $\valueQPraw{\vcost}{\gamma}$ and policy evaluation 
$\valueVPraw{\vcost}{\gamma}$ for policy $\vPiP$ for the 
$\mathcal{M'} = \langle S, A, T, \costC, G, \sched' \rangle$ where
$\sched' = (\kappa, \sched(0), \sched(1),\dots)$
is:
\begin{align*}
\valueVP{\vcost}{s}{\vPiP}{\sched'}{\gamma}\,  = & \;\;\vcost(s, \vPiP(s)) \quad + \\
&\hspace{5ex} \gamma\!\!\sum_{s' \in S}{\vT(s', \vPiP(s), s) \valueV{\vcost}{s'}{\vPi}{\sched}{\gamma}},\\[-28pt]
\end{align*}
and 

\vspace*{-18pt}
\begin{align*}
&\valueQP{\vcost}{s}{\vv{a}}{\sched'}{\gamma}\,  =  \;\;\vcost(s, \vv{a}) \quad + \\
&\hspace{9ex} \gamma\!\!\sum_{s' \in S}{\vT(s', \vv{a}, s)\, \valueQ{\vcost}{s'}{\vPi(s', 0)}{\sched}{\gamma}}, \nonumber
\label{eq:value-fxn}
\end{align*}
where $\vPiP(s, n) = \vPi(s,n-1)$, for $n\geq 1$, and $\vPiP(s, 0)$ is obtained from $\min\limits_{\vv{a}\in A^\kappa}
\valueQP{\vcost}{s}{\vv{a}}{\sched'}{\gamma}$.
\end{proposition}

\begin{proofsketch}
The computation here is reminiscent of value iteration. 
In fact, if $\sched = \seq{\recc{\kappa}}$, the $\valueQPraw{\vcost}{\gamma}$ update is equivalent to a single pass of value iteration, which makes sense considering $\sched^\prime = \seq{\kappa, \sched(0), \sched(1),\dots} = \seq{\kappa, \recc{\kappa}} = \seq{\recc{\kappa}} = \sched$. 
When this is not the case, however, the key difference is the new values computed, $\valueQPraw{\vcost}{\gamma}$, are separate from the old values used, $\valueQraw{\vcost}{\gamma}$. 
One can think of this separation in terms of parallel, stacked \emph{layers} of value functions. When an agent takes the first step with stride $\sched^\prime(0) = \kappa$, it moves from a state $s$ to a state $s^\prime$ and also moves vertically from the layer of $\sched^\prime$ to the lower layer of $\sched$, where it repeats with shorter and shorter schedules. 
The value starting from $s$ is simply the expected cost from moving, $\vcost(s, \vv{a})$, plus the value of the shorter schedule continuing from $s^\prime$, represented by the lower layer. 
The same argument applies for the policy evaluation. Indeed, the only difference between the two is that the value function selects the minimum cost action while the policy evaluation selects the policy action.
%
\end{proofsketch}

The fact that schedules may, thus, be constructed from those of shorter length
is the basis of the approach---the overall algorithm for which is presented in
Algorithm \ref{alg:with-filtering}.  Indeed, every schedule in the search space
can be recursively constructed from shorter ones, all the way down to the
$\strideBound$ base cases of recurring tail schedules $\recc{\kappa}$.  
As shown in lines 4--9
of Algorithm~\ref{alg:with-filtering}, these base \psmdps can each be represented as
a periodically state-observed Markov Decision Process (\psomdp) introduced in
\cite{rossi22periodic} and solved as ordinary MDPs, yielding a set of converged
value functions for use as a base for schedule extensions.

Extending the length of a schedule by adding a prefix requires only a single
pass; the key schedule extension procedure appears in
Algorithm~\ref{alg:schedule-extension}.  The overlapping substructure 
allows us to apply a dynamic programming approach to constructing
schedules, demonstrated in lines 13--18 of Algorithm~\ref{alg:with-filtering},
re-using values for already constructed schedules that make up each suffix
rather than computing them anew.

\subsection{Filtering}
The recursive construction process markedly reduces the work done per schedule, but the overall complexity remains significant (exponential). As schedule lengths grow, a mechanism is needed to reduce the number of candidate schedules to extend. By filtering out schedules, fewer reach the final stage of Pareto front calculations, resulting in faster computation but potentially lower solution quality. As such, care must be taken in how schedules are filtered out. 

One approach would be to pick only the best schedules at each stage, using the same criteria as the final Pareto front: the initial distribution's expected execution and check-in costs for each schedule. 
This approach, however, fails as a poor schedule with respect to the initial distribution may result in a good one once extended and vice versa. 
(We explore this as a case study later in Section~\ref{subsubsec:splitter}.)

To mitigate this issue, we turn to a different evaluation criteria when filtering. Schedule suffixes do not start from the initial distribution and, as such, their policy values are  not accurately represented when evaluated against the initial distribution; to alleviate this, we examine them against the other states as well. 
One such method is to use the uniform distribution, i.e., taking the average of the costs over every state.
Rather than selecting schedules that work well starting in a limited number of states, we select schedules that work well on average. This filtering is done at the end of every extension stage, as seen in lines 19--26 of Algorithm~\ref{alg:with-filtering}.

We further tune the aggressiveness of the filtering with a \textit{margin} parameter that also keeps schedules whose policy values are within a prescribed distance from the Pareto front, rather than just the ones on the front.
Another option is to use multiple independent distributions and be able to favor certain states like the greedy approach,  while still having widespread support. 
Each distribution provides two additional dimensions in the Pareto front,
resulting in less aggressive filtering. 
Taken to the extreme, one can select the collection of $|S|$ distributions with a Dirac delta distribution over each and every state, but in most cases not a single schedule will be dominated in the higher dimensional Pareto front and, as a result, none will be filtered out.

\begin{algorithm}[H]
\scriptsize
{\label{pro:paretoFrontSchedules}\functype{ParetoFrontSchedules}$($

$\quad \quad \quad \mathcal{M}$, \rm{the base MDP representing the environment}

$\quad \quad \quad \argtype{strides}$, \rm{the set of available check-in periods}

$\quad \quad \quad \argtype{length}$, \rm{the maximum length schedule to explore}

$\quad \quad \quad \mathbf{\InitD}$, \rm{initial distribution of start states}

$\quad \quad \quad \argtype{filter}$, \rm{whether to filter out schedules during building}

$\quad \quad \quad \argtype{distribs}$, \rm{the set of distributions to use in filtering}

$\quad \quad \quad \argtype{margin}$, \rm{distance from front to keep during filtering}

$\quad \quad \quad \argtype{alphas}$, \rm{the alpha values for intermediary policy computation})
}

    \begin{algorithmic}[1]
    \State{$\vartype{allSchedules} \leftarrow \{\}$} \RightComment{All candidate schedules}
    \State{$\vartype{stages}[0] \leftarrow \{\}$} \RightComment{Sets for schedules of each length}

    \LeftComment{0}{Construct initial $\recc{k}$ schedules}
    \State{\algorithmicforall $~k \in \vartype{strides}$ \textbf{do}}  

    \LeftComment{1}{Construct composite action process MDP with $k$-length macro-action sequences}
    \State {$\quad \mathcal{M}_k \leftarrow {\functype{CompositeMdp}}(\mathcal{M}, k)$} \RightComment{See Composite Action Process in \cite{rossi22periodic}}
    \State {$\quad \vcostK(s, \vv{a}) \leftarrow \vcostC\left(s,\vv{a}\right)$ with action costs set to 1 as per Eq. (\ref{ref:cost})} 
    \LeftComment{1}{Policy evaluations for $\costOptPiRaw{\exec}$ and $\costOptPiRaw{\comms}$: $\valueV{\vcostC}{\cdot}{\cdot}{\sched}{\gamC}$ for execution cost and $\valueV{\vcostK}{\cdot}{\cdot}{\sched}{\gamK}$ for check-in cost (see \ref{subsec:dominate})}
    \State {$\quad \valueV{\vcostC}{\cdot}{\costOptPi{\exec}}{\sched}{\gamC} \leftarrow $ solve $\mathcal{M}_k$ w.r.t. $\vcostC$ for $\costOptPi{\exec}$}
    \State {$\quad \valueV{\vcostK}{\cdot}{\costOptPi{\exec}}{\sched}{\gamK} \leftarrow {\functype{PolicyEval}}(\vcostK, \mathcal{M}_k, \costOptPi{\exec})$}
    
    \State {$\quad \valueV{\vcostK}{\cdot}{\costOptPi{\comms}}{\sched}{\gamK} \leftarrow $ solve $\mathcal{M}_k$ w.r.t. $\vcostK$ for $\costOptPi{\comms}$}
    
    \State {$\quad \valueV{\vcostC}{\cdot}{\costOptPi{\comms}}{\sched}{\gamK} \leftarrow {\functype{PolicyEval}}(\vcostC, \mathcal{M}_k, \costOptPi{\comms})$}
    \State {$\quad \sched \leftarrow \classtype{Schedule}(\paramtype{Strides}$ = $[k], \valueV{\vcostC}{\cdot}{\cdot}{\sched}{\gamC}, \valueV{\vcostK}{\cdot}{\cdot}{\sched}{\gamK})$}
    \State {$\quad \vartype{stages}[0]$.append($\sched$)}
    \State {$\quad \vartype{allSchedules}$.append($\sched$)}

    \LeftComment{0}{Extending schedules}
    \State{\algorithmicfor $~i \leftarrow 1$ to ($\vartype{length}-1)$ \textbf{do}}
    \State {$\quad \vartype{stages}[i] \leftarrow \{\}$}
    \State{\quad \algorithmicforall $~(\sched,k) \in \vartype{stages}[i-1] \times \vartype{strides} $ \textbf{do}}
    \LeftComment{2}{Add $k$ to head of schedule}
    \State {$\quad \quad \sched^\prime \leftarrow $\functype{PrependSchedule}$(\sched, k, \mathcal{M}_k, \vcostC, \vcostK, \vartype{alphas})$} 
    \State {$\quad \quad \vartype{stages} [i]$.append($\sched^\prime$)}
    \State {$\quad \quad \vartype{allSchedules}$.append($\sched^\prime$)}
    \textcolor{\filteringColor}{
        \LeftComment{1}{Filtering}
        \State{\quad \textbf{if} filtering \textbf{then}}
        \State{\quad \quad \algorithmicforall $~(\sched, d) \in \vartype{allSchedules} \times \vartype{distribs}$}
        \State{\quad \quad \quad $\executionCost{\sched}{\cdot}_d = \valueV{\vcostC}{d}{\cdot}{\sched}{\gamC}$ and $\checkinCost{\sched}{\cdot}_d = \valueV{\vcostK}{d}{\cdot}{\sched}{\gamK}$}
        \State {$\quad \quad P_R \leftarrow \functype{ParetoFront}($all realizable fronts$)$} \RightComment{see \ref{subsec:bounds} on Pareto fronts}
        \State {$\quad \quad P \leftarrow $ all $\sched \in \vartype{allSchedules}$ with optimistic front not dominated by $P_R$}
        \State {$\quad \quad M \leftarrow$ all $\sched \in \vartype{allSchedules}$ with optimistic front within \vartype{margin} distance of $P_R$}
        \State {$\quad \quad \vartype{allSchedules} \leftarrow P \cup M$}
        \State {$\quad \quad \vartype{stages}[i] \leftarrow \vartype{stages}[i] \cap \vartype{allSchedules}$}
    }

    \State{\algorithmicforall $~\sched \in \vartype{allSchedules}$ \textbf{do}} 
    \State{\quad $\executionCost{\sched}{\cdot} = \valueV{\vcostC}{\InitD}{\cdot}{\sched}{\gamC}$ and $\checkinCost{\sched}{\cdot} = \valueV{\vcostK}{\InitD}{\cdot}{\sched}{\gamK}$}
    \State {$P_R \leftarrow \functype{ParetoFront}($all realizable fronts$)$}
    \State {$P \leftarrow $ all $\sched \in \vartype{allSchedules}$ with optimistic front not dominated by $P_R$}

    \State {\textbf{return} $P$}

    \end{algorithmic}
    \caption{Algorithm with filtering.}
    \label{alg:with-filtering}
\end{algorithm}

\begin{algorithm}[h]
\scriptsize
{\label{pro:MDP2LP}\functype{PrependSchedule}$($

$\quad \quad \quad \sched$, \rm{the} \classtype{Schedule} \rm{to extend}

$\quad \quad \quad k$, \rm{the stride to insert at the schedule head}}

$\quad \quad \quad \mathcal{M}_k$, \rm{the composite MDP representing $k$-length macro actions}


$\quad \quad \quad \vcostC, \vcostK$, \rm{the macro execution and check-in cost functions, respectively}

$\quad \quad \quad \argtype{alphas}$, \rm{the alpha values for intermediary policy computation})

    \begin{algorithmic}[1]
    \State{$\sched^\prime \leftarrow$ copy of $\sched$} \RightComment{Below, $\functype{ExtPolicyEval}$ means extend policy evaluation}
    \State{$\sched^\prime.\paramtype{Strides}$.insert(0, $k$)} \RightComment{Insert $k$ as first element}
    
    \LeftComment{0}{Execution cost of $\costOptPiRaw{\exec}$ is value function on $\vcostC$ while check-in cost is evaluation of $\costOptPiRaw{\exec}$ on $\vcostK$.}
    \State{$\valueV{\vcostC}{\cdot}{\costOptPi{\exec}^\prime}{\sched^\prime}{\gamC} \leftarrow \functype{ExtendValueFunction}(\valueV{\vcostC}{\cdot}{\costOptPi{\exec}}{\sched}{\gamC}, \mathcal{M}_k)$}
    \State{$\valueV{\vcostK}{\cdot}{\costOptPi{\exec}^\prime}{\sched^\prime}{\gamK} \leftarrow \functype{ExtPolicyEval}(\valueV{\vcostK}{\cdot}{\costOptPi{\exec}}{\sched}{\gamK}, \mathcal{M}_k, \costOptPi{\exec}^\prime)$}

    \LeftComment{0}{Check-in cost of $\costOptPiRaw{\comms}$ is value function on $\vcostK$ while execution cost is evaluation of $\costOptPiRaw{\comms}$ on $\vcostC$}

    \State{$\valueV{\vcostK}{\cdot}{\costOptPi{\comms}^\prime}{\sched^\prime}{\gamK} \leftarrow \functype{ExtendValueFunction}(\valueV{\vcostK}{\cdot}{\costOptPi{\comms}}{\sched}{\gamK}, \mathcal{M}_k)$}
    \State{$\valueV{\vcostC}{\cdot}{\costOptPi{\comms}^\prime}{\sched^\prime}{\gamC} \leftarrow \functype{ExtPolicyEval}(\valueV{\vcostC}{\cdot}{\costOptPi{\comms}}{\sched}{\gamC}, \mathcal{M}_k, \costOptPi{\comms}^\prime)$}


    \State{\algorithmicfor \textbf{ each} $\alphalabel \in \vartype{alphas}$ \textbf{do}}

    \State{$\quad U_\alphalabel \leftarrow \alphalabel \times \valueV{\vcostC}{\cdot}{\costOptPi{\exec}^\prime}{\sched^\prime}{\gamC} + (1-\alphalabel) \times \valueV{\vcostK}{\cdot}{\costOptPi{\comms}^\prime}{\sched^\prime}{\gamK}$}
    \State{$\quad \vcostA \leftarrow \alphalabel \times \vcostC + (1-\alphalabel) \times \vcostK$}
    \State{$\quad \costOptPi{\alphalabel}^\prime \leftarrow$ policy from $ U_\alphalabel$, $\vcostA$, and $\mathcal{M}_k$}

    \State{$\quad \valueV{\vcostC}{\cdot}{\costOptPi{\alphalabel}^\prime}{\sched^\prime}{\gamC} \leftarrow \functype{ExtPolicyEval}(\valueV{\vcostC}{\cdot}{\costOptPi{\alphalabel}}{\sched}{\gamC}, \mathcal{M}_k, \costOptPi{\alphalabel}^\prime)$}
    \State{$\quad \valueV{\vcostK}{\cdot}{\costOptPi{\alphalabel}^\prime}{\sched^\prime}{\gamK} \leftarrow \functype{ExtPolicyEval}(\valueV{\vcostK}{\cdot}{\costOptPi{\alphalabel}}{\sched}{\gamK}, \mathcal{M}_k, \costOptPi{\alphalabel}^\prime)$}
    

    \State {\textbf{return} $\sched^\prime$}
    \end{algorithmic}
    \caption{Schedule extension procedure.}
    \label{alg:schedule-extension}
\end{algorithm}

\vspace{-.5em}
\section{Algorithm Characterization} \label{sec:character}

\subsection{Case Studies}
We provide two case studies to better illustrate the reasoning behind the algorithm. 

\subsubsection{Corridor grid with two cadences in sequence}

Our main case study is a grid world featuring a series of walls in sequence, directly motivated by the example in \cite[Sect.~III-B]{rossi22periodic} showing that a higher frequency of check-ins is not always better. The central idea is that columns of walls are placed with a certain cadence.

\begin{figure}[h]
    \includegraphics[width=\linewidth]{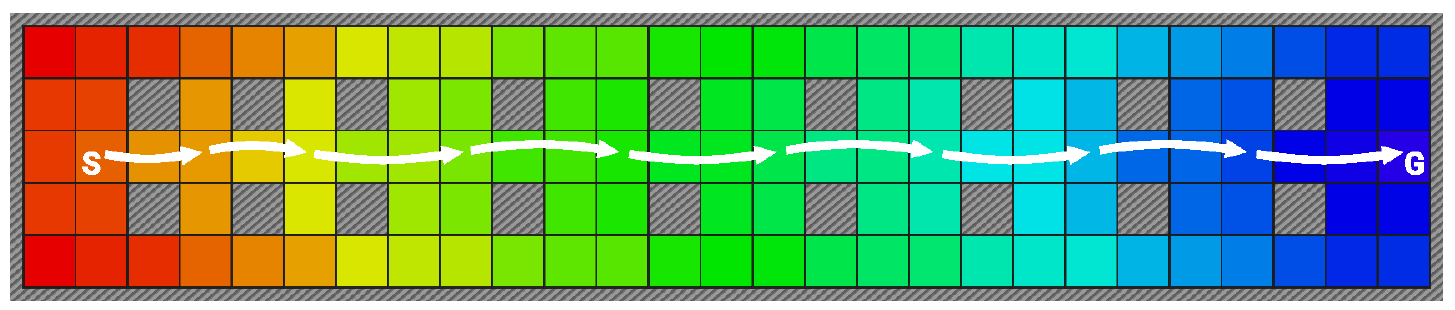}
    \caption{The $\pi^\opt$ policy for $22\recc{3}$ schedule in corridor grid world.
    \label{fig:corridor_world}}
\end{figure}

A cadence of 3 means that it takes 3 horizontal steps to reach one wall from another. The agent chooses between moving in a cardinal direction or waiting (a ``no-op" action). Movement has a chance of causing an additional drift left or right of the movement direction, while no-ops incur no drifting. 
As a result, the agent prefers a check-in immediately before each wall column to verify it has not drifted vertically. 
The $\pi^\opt$ policy, thus, has the agent wait for a check-in when it believes it has reached the preferred spot, rather than risk moving when the next check-in is still in the future. 
This behavior leads to an interesting phenomenon, where a stride that matches the environment cadence well 
will lead to fewer waiting delays than a more frequent check-in period.

\begin{SCfigure}
    \includegraphics[width=0.65\linewidth]{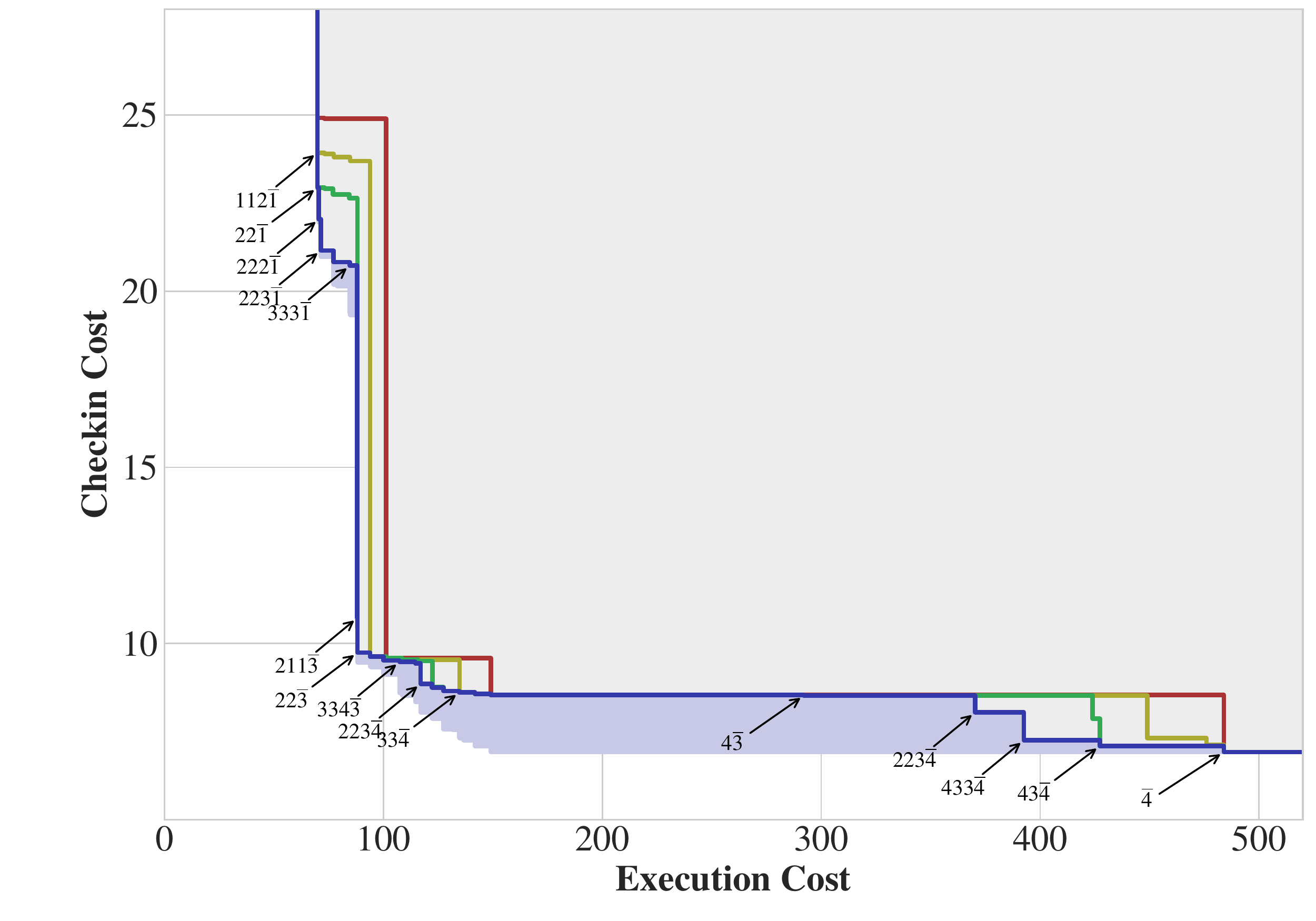}
    \caption{Progression of the Pareto front (red, yellow, green, blue) as schedules get longer, up to length 4.\label{fig:pareto_c4_lengths}\newline}
\end{SCfigure}

\begin{figure*}
\centering
    \begin{tabular}{@{}c@{}c@{}c@{}c@{}c@{}c@{}c@{}c@{}c@{}c@{}c@{}}
     \includegraphics[width=\xAcross]{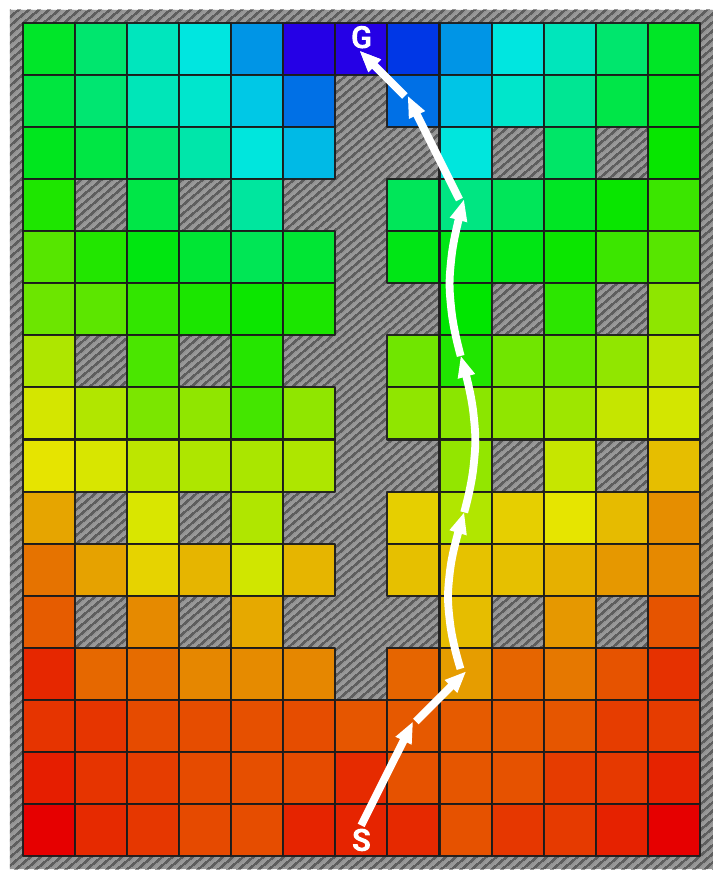} &   
     \includegraphics[width=\xAcross]{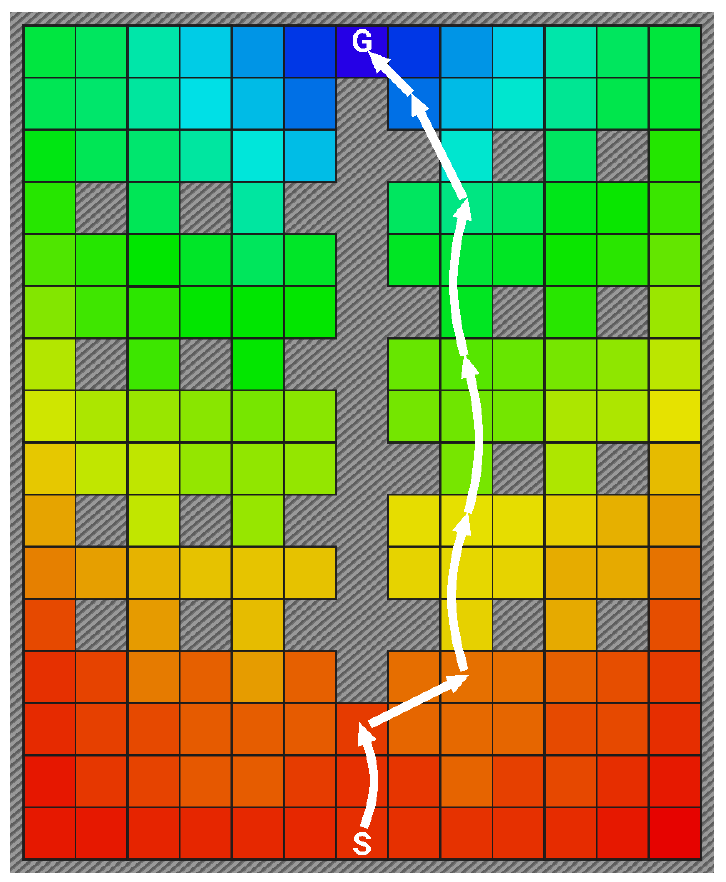} &
     \includegraphics[width=\xAcross]{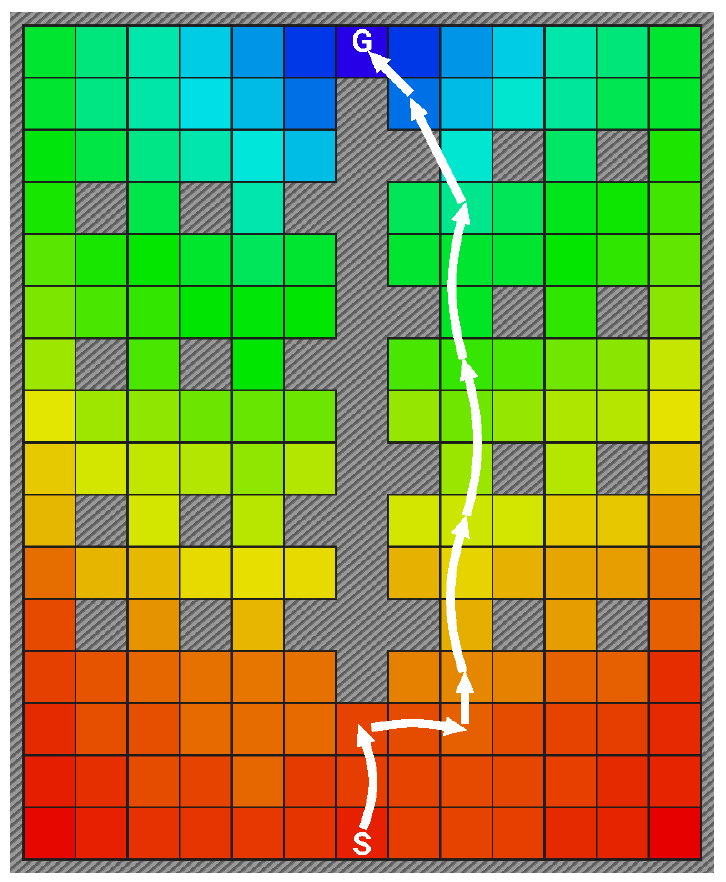} &   
     \includegraphics[width=\xAcross]{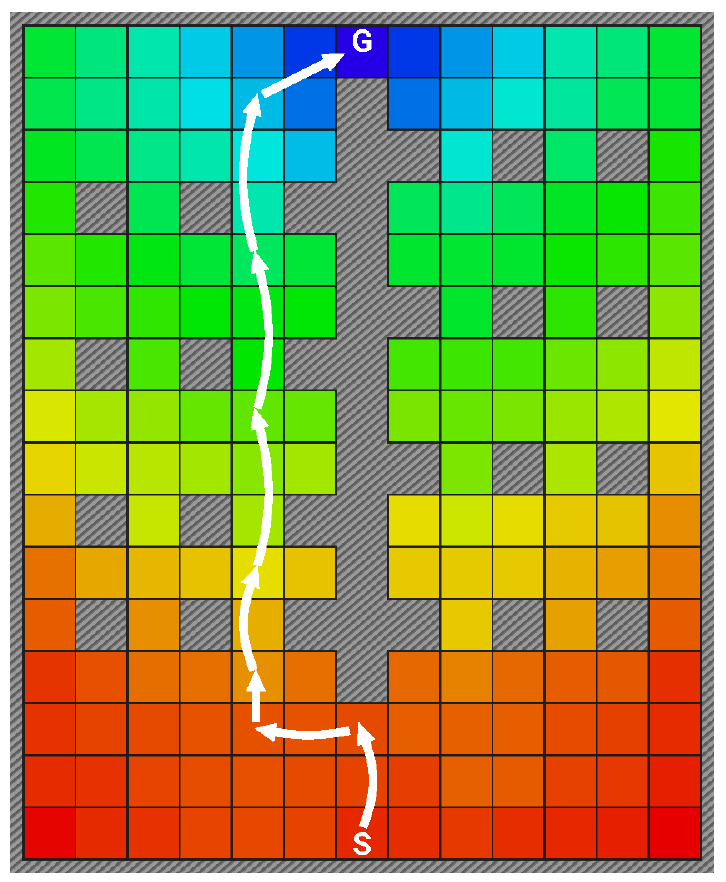} &
     \includegraphics[width=\xAcross]{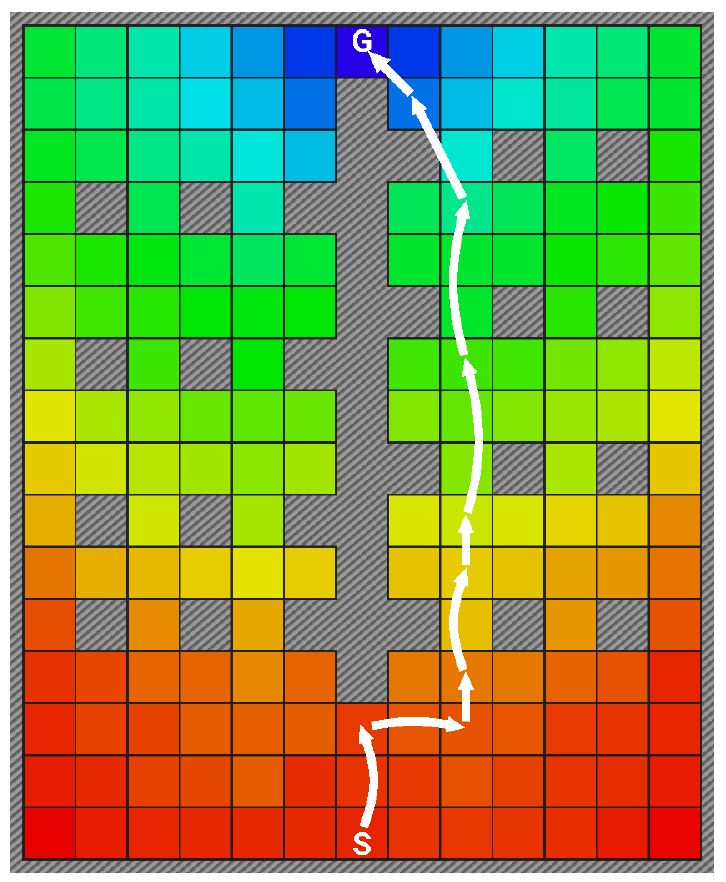} &   
     \includegraphics[width=\xAcross]{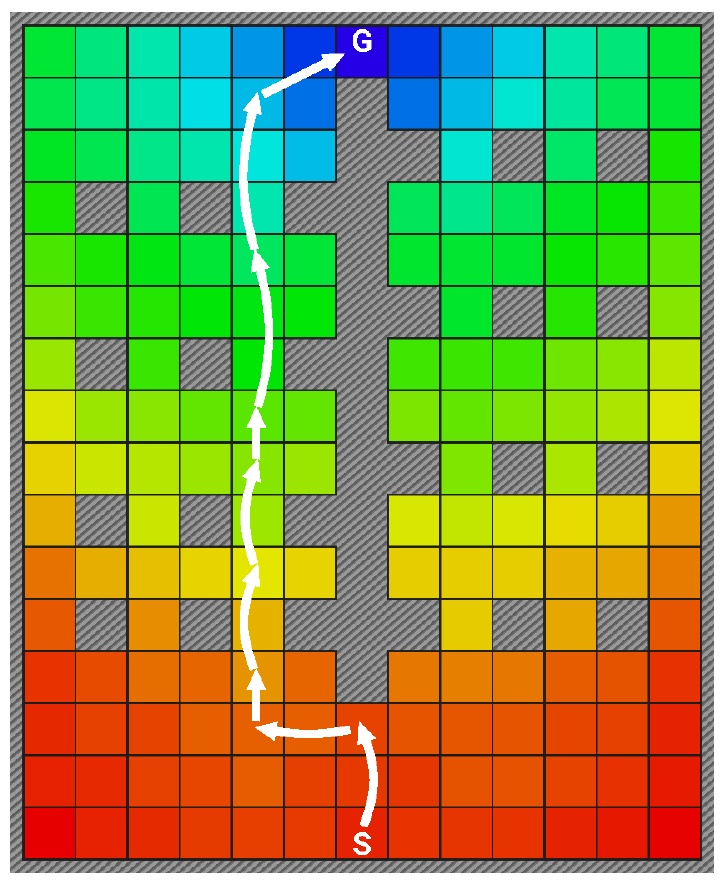} & 
     \includegraphics[width=\xAcross]{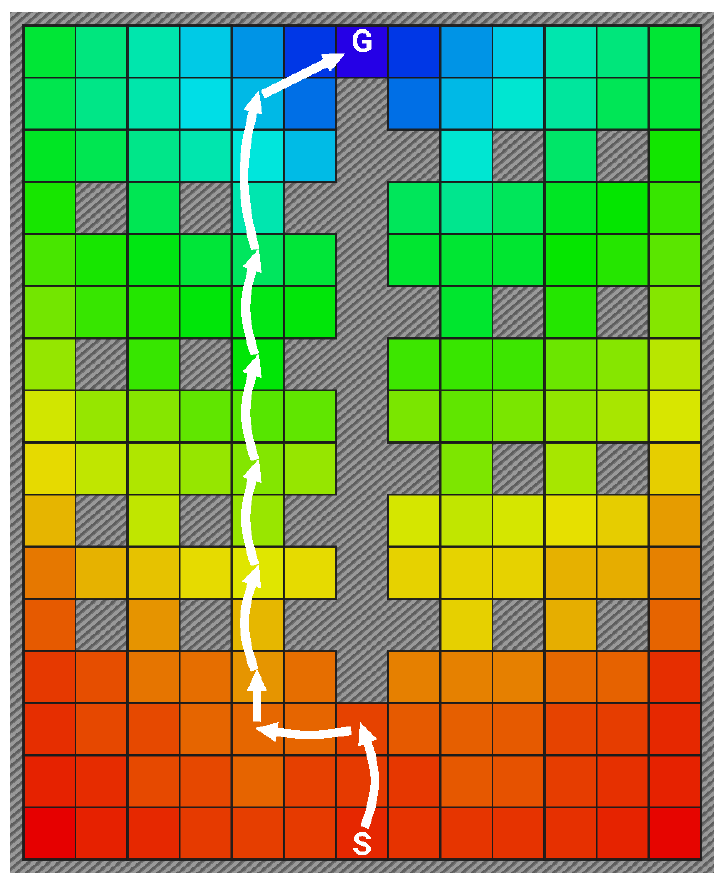} & 
     \includegraphics[width=\xAcross]{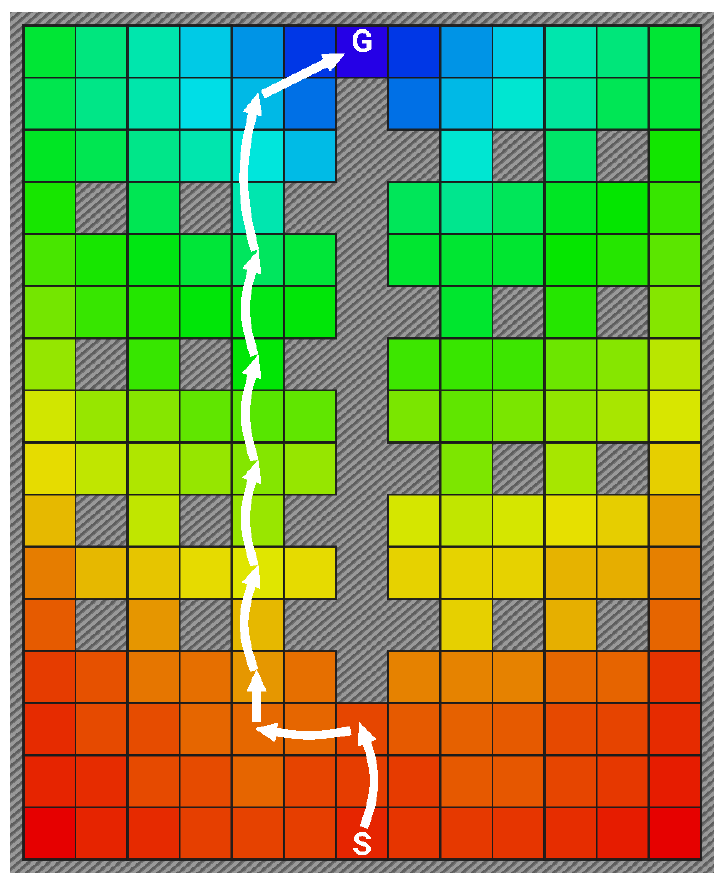} & 
     \includegraphics[width=\xAcross]{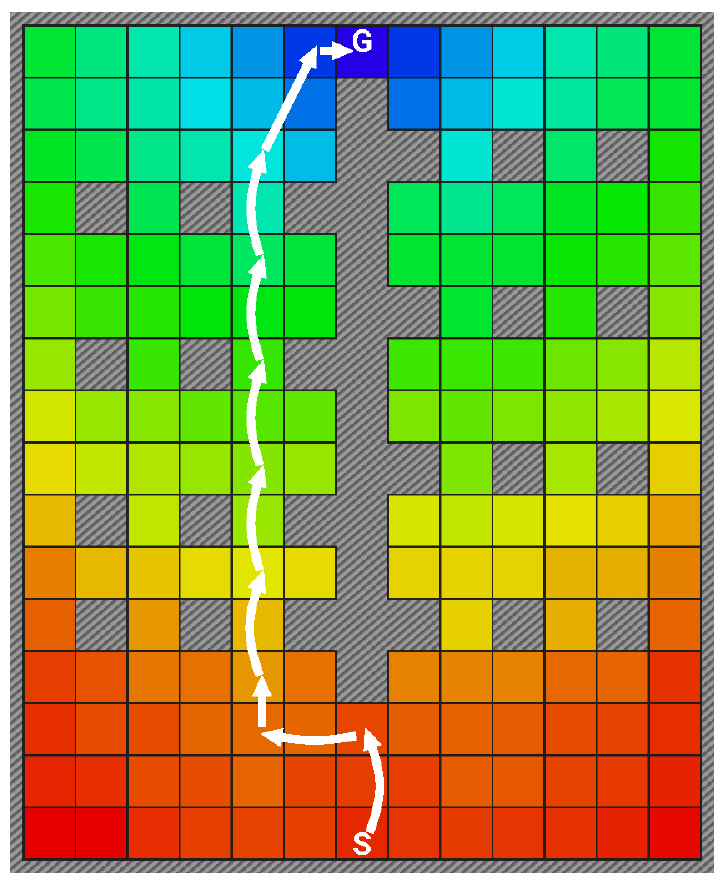} & 
     \includegraphics[width=\xAcross]{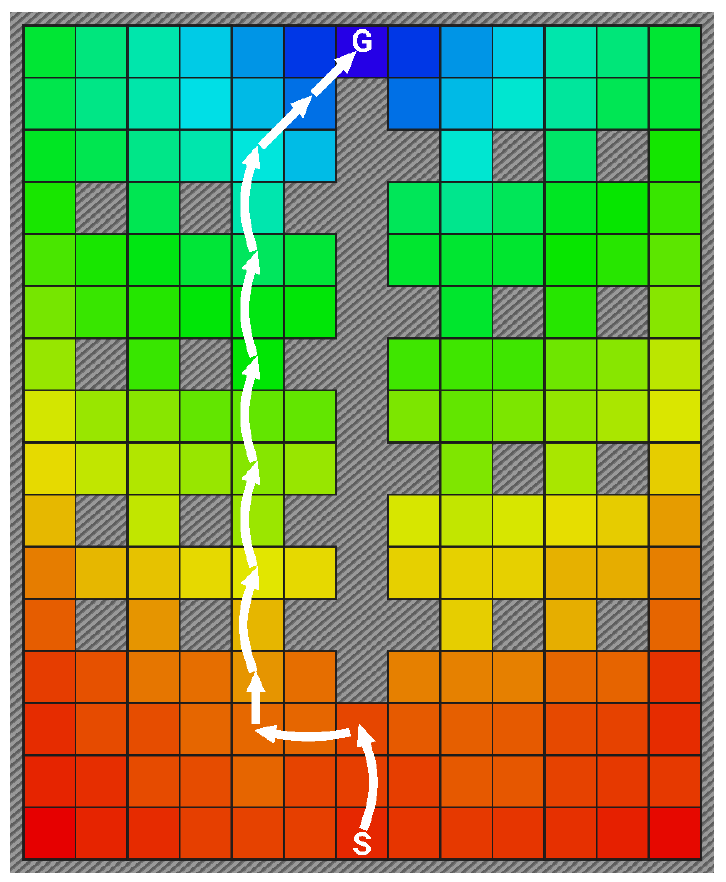} \\
    \footnotesize \bf{\textsf{(a)}}\;\recc{3} & 
    \footnotesize \bf{\textsf{(b)}}\;$2\recc{3}$ & 
    \footnotesize \bf{\textsf{(c)}}\;$22\recc{3}$ & 
    \footnotesize \bf{\textsf{(d)}}\;$222\recc{3}$ & 
    \footnotesize \bf{\textsf{(e)}}\;$2222\recc{3}$ & 
    \footnotesize \bf{\textsf{(f)}}\;$22222\recc{3}$ & 
    \footnotesize \bf{\textsf{(g)}}\;$2^6\recc{3}$ & 
    \footnotesize \bf{\textsf{(h)}}\;$2^7\recc{3}$ & 
    \footnotesize \bf{\textsf{(i)}}\;$2^8\recc{3}$ & 
    \footnotesize \bf{\textsf{(j)}}\;$2^9\recc{3}$ 
    \end{tabular}
    \caption{Optimal policies may switch between going left and going right when schedules are extended.\label{fig:splitter_world}}
\end{figure*}

\begin{figure*}[b]
    \vspace{1em}
    \centering
    \begin{subfigure}{.32\linewidth}
        \centering
        \runningTimePlot{plots/running_tests-uniform.csv}
        \vspace{-1.5em}
        \caption{Uniform distribution\label{fig:running_time-uniform-c4l4}}
    \end{subfigure}
        \hfill
    \begin{subfigure}{.32\linewidth}
        \centering
        \runningTimePlot{plots/running_tests-initial.csv}
        \vspace{-1.5em}
        \caption{Initial distribution\label{fig:running_time-initial}}
    \end{subfigure}
    \hfill
    \begin{subfigure}{.32\linewidth}
        \centering
        \runningTimePlot{plots/running_tests-mixed.csv}
        \vspace{-1.5em}
        \caption{Both uniform and initial distributions
        \label{fig:running_time-mix}}
    \end{subfigure}
    \caption{Planning time and precision \vs margin for check-in periods 1 to 4, up to schedule length 4 for different filtering distributions.}\label{fig:running_times_dist}
\end{figure*}

One would expect that the optimal schedule would 
mirror the cadences: $22\recc{3}$. Figure~\ref{fig:corridor_world} demonstrates how each stride 
moves to the spot before each column. Indeed, this schedule appears in the Pareto front shown in Figure~\ref{fig:pareto_c4_lengths}, dominating both $\recc{2}$ and $\recc{3}$, and would be the optimal schedule if both check-in cost and execution cost were valued equally. 


The Pareto fronts in Figure~\ref{fig:pareto_c4_lengths} correspond to the different stages of schedule extension. 
As schedules are extended, they form new fronts with existing schedules, progressing towards the final front for all schedules up to length 4. 
Note how most of the base schedules---$\recc{1}$, $\recc{2}$, and $\recc{3}$---never make it to the final front, yet appear as suffixes in those that do. 
They are sub-optimal candidates by themselves from the start state, but the prefixes bring the agent to an optimal state for the suffixes to continue
, which is why it is important to evaluate the suffixes in states apart from the start when filtering. 

The corridor world has a recurring goal reward of \num{10000}, movement cost of \num{1}, zero no-op cost, collision cost of \num{300000}, drift probabilities of \SI{5}{\percent} left and \SI{5}{\percent} right of the intended direction, and a base discount factor of $\sqrt{0.99}$.

\subsubsection{Splitter grid with two cadences in parallel}  \label{subsubsec:splitter}


We further illustrate how schedule extension can result in non-trivial changes to the policy of a schedule, with the grid world in Figure~\ref{fig:splitter_world}. 
The world is constructed such that a $\recc{2}$ schedule would favor the west side due to the cadence 2 rows, while $\recc{3}$ would favor the east.
Due to the center column, the agent cannot switch once it takes a side. 
At the start state, the choice is clear: go left if the stride is 2, go right if the stride is 3. 
As we start prepending $k=2$ check-ins to the $\recc{3}$ schedule, however, we notice that the policy flips directions for certain schedules, with $222\recc{3}$ going leftwards instead. This occurs since the prefix 
 of the schedule, $222$, brings the agent to states on the left that are superior for the remainder of the schedule, $\recc{3}$, than if it had gone right. As we continue prepending more $k=2$ check-ins, the flipping continues as the tail gets pushed farther north from the start state.


\subsection{Time Complexity}
The standard procedure for solving MDPs, value iteration, has time complexity $O({|S|}^2 {|A|})$ per iteration but requires multiple iterations to converge. If we assume some $I$ iterations to converge, the complexity for solving each base \psmdp is $O({I |S|}^2 {|A|^{\strideBound}})$, as there are $|A|^{\strideBound}$ actions in the composite MDP. 
Since our approach re-uses already computed schedules, it requires only the $\strideBound$ base convergences and $\strideBound^n - \strideBound$ extensions of a single pass per alpha. If we account for the proportion $f$ of schedules filtered out, we have our final complexity of $O(|\alpha|(\strideBound I + (1-f) \strideBound^n) ({|S|}^2 {|A|}^{\strideBound}))$. 
In our experimental results, we have cases of $f$ ranging from 0.12 (high margins) to 0.993 (low margins plus alphas). 

\section{Results} \label{sec:results}


\subsection{Margin and Distribution Effectiveness}
Figure~\ref{fig:running_times_dist} shows the effect of the margin parameter on running time, quality, and filtering aggressiveness on the corridor case study. The quality metric is obtained by comparing the final Pareto front with the true front generated without filtering. A margin value of zero results in the most aggressive pruning, keeping only non-dominated schedules on each stage's Pareto front. As a result, it has the lowest running time and solution quality, both of which steadily increase with higher margin values as filtering decreases. 
Some distributions are better suited than others in certain aspects: Figure~\ref{fig:running_time-initial} shows how 
filtering with the initial distribution $\InitD$ starts out with both a slightly higher solution quality and a lower running time than its uniform distribution counterpart in Figure~\ref{fig:running_time-uniform-c4l4}, but is much less effective in raising quality by increasing the margin. This is a result of the greedy pitfall: it succeeds at recognizing suffixes optimal from the start state, but fails to capture schedules with suffixes only optimal from a different state. It then requires a higher margin tolerance to avoid filtering the latter case out early. Figure~\ref{fig:running_time-mix} shows how using both distributions in parallel in a multi-dimensional Pareto front appears to balance both. The initial distribution ensures that the easy cases are captured, while the uniform distribution prevents the harder cases from being dropped---ultimately, it gives improved quality even without a margin.

\begin{figure}[h]
    \centering
    \runningTimePlotLen{plots/running_tests-truthVmixed-combined.csv}
    \caption{Planning time and precision \vs problem size, with alphas \{0.2, 0.4, 0.6, 0.8\} (dashed) and without (solid). Planning time normalized against running time without filtering.\label{fig:running_time-uniform-sizes}}
\end{figure}

\blockcomment{
\begin{figure}[t]
     \centering
        \runningTimePlot{plots/running_tests-uniform.csv}
        \caption{Planning time and precision \vs margin tolerance for check-in periods 1 to 4, up to sequence length 4 using uniform distribution for filtering.\label{fig:running_time-uniform-c4l4}}
\end{figure}

\begin{figure}[t]
     \centering
        \runningTimePlot{plots/running_tests-initial.csv}
        \caption{Planning time and precision \vs margin tolerance for check-in periods 1 to 4, up to sequence length 4 using initial distribution for filtering.\label{fig:running_time-initial}}
\end{figure}

\begin{figure}[t]
     \centering
        \runningTimePlot{plots/running_tests-mixed.csv}
        \caption{Planning time and precision \vs margin tolerance for check-in periods 1 to 4, up to sequence length 4 using both uniform and initial distributions for filtering.\label{fig:running_time-mix}}
\end{figure}

\begin{figure}[t]
     \centering
        \runningTimePlotLen{plots/running_tests-truthVmixed-combined.csv}
        \caption{Planning time and precision \vs problem size, without midpoints (solid) and with midpoints \{0.2, 0.4, 0.6, 0.8\} (dashed). Filtered using mixed distribution with no margin tolerance. Running times normalized against running time without filtering.\label{fig:running_time-uniform-sizes}}
\end{figure}
}

\subsection{Scaling}
Figure~\ref{fig:running_time-uniform-sizes} summarizes different problem sizes to demonstrates how filtering can significantly mitigate exponential growth of the search space. 
One can easily see how increasing schedule lengths rapidly becomes a problem with $\strideBound^n$ schedules: 
for $n = 9$ the unfiltered algorithm took 15 hours, compared to 66 seconds for $n = 4$. 
As the problem size grows and filtering is able to trim out more schedules, we see that avoiding extending sub-optimal candidates improves the running time significantly without sacrificing precision. 

\begin{SCfigure}
    \includegraphics[width=0.66\linewidth]{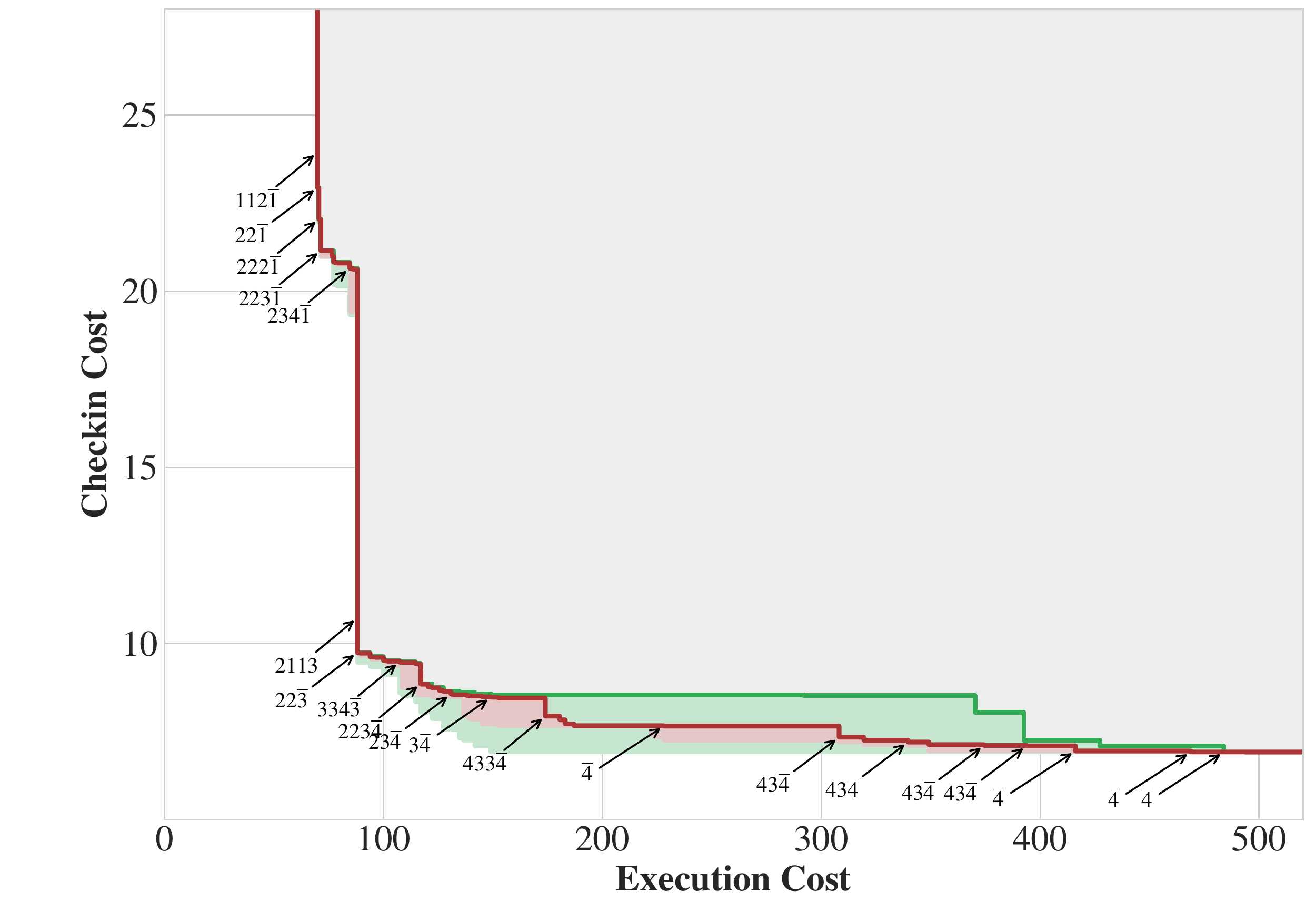}
    \caption{Pareto fronts with alphas \{0.1, 0.2, 0.3, 0.4, ..., 0.9\} (red) and without (green) at length 4, with shaded regions between the optimistic fronts and the bold realizable fronts.\label{fig:pareto_c4_alphas}}
\end{SCfigure}

We also see an interesting phenomenon appear in the introduction of alpha values. They increase running time for shorter schedules as one would expect, due to needing to compute the extra policies, but also allow filtering to be more effective at dismissing schedules from consideration, by reducing the region of uncertainty that each schedule occupies. This performance benefit is not enough to overcome the overhead from extra computation for shorter schedules, but when asking for longer schedules, the savings from increased trimming scales and reduces the computation time overall, bringing the filtered running time down from 90 minutes to 36 for $n=9$. Their effect on Pareto fronts is also clearly visible in Figure~\ref{fig:pareto_c4_alphas}. The green region is derived without alpha values---that is, with only 
$\costOptPi{\exec}$ and
$\costOptPi{\comms}$ for each schedule. The red region uses 9 additional policies per schedule, resulting in much tighter realizable and optimistic fronts.

\begin{figure}[t]
    \centering
    \begin{subfigure}{.45\linewidth}
        \centering
        \includegraphics[width=\textwidth]{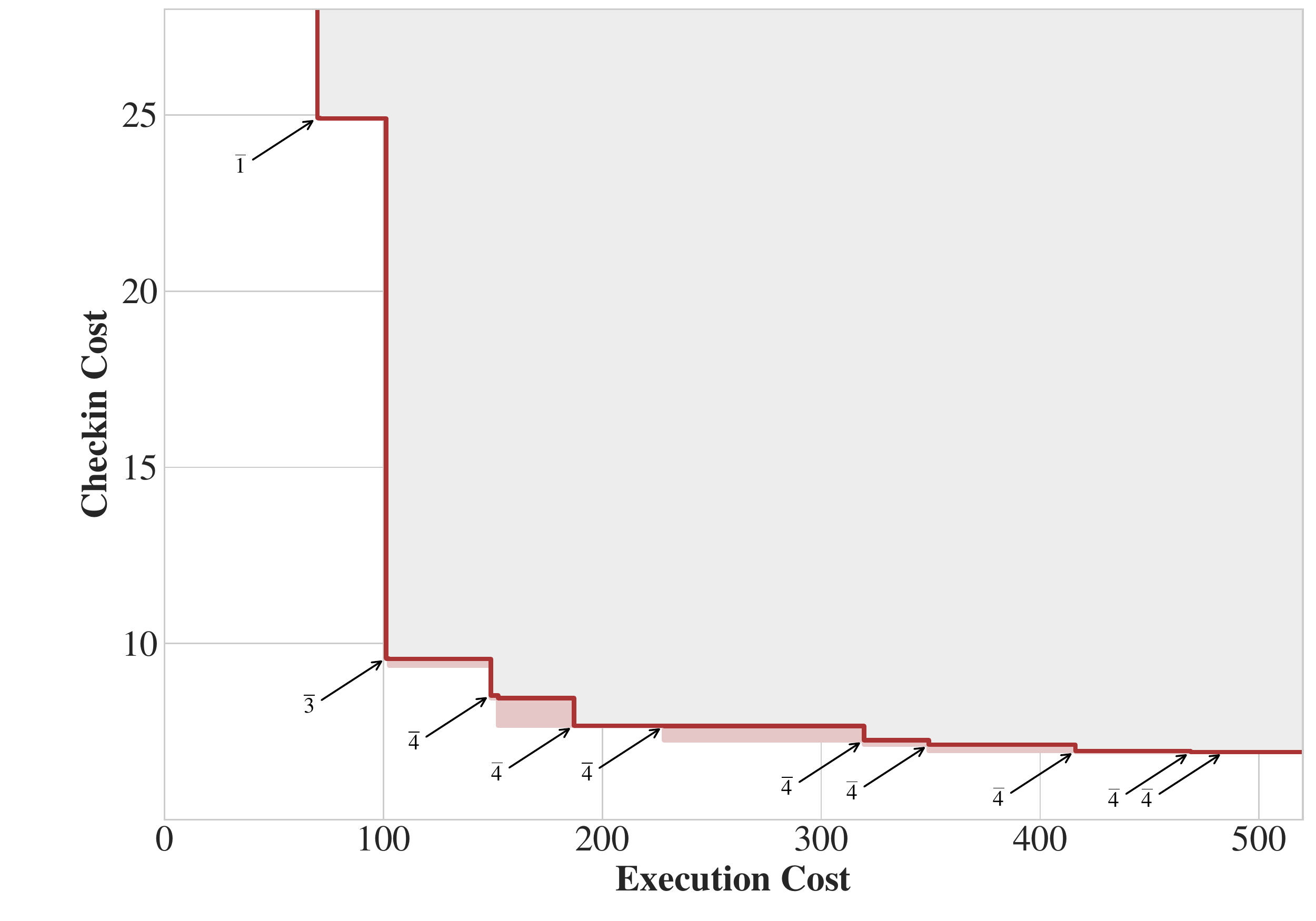}
        \caption{Up to length 1}
    \end{subfigure}
        \hfill
    \begin{subfigure}{.45\linewidth}
        \includegraphics[width=\textwidth]{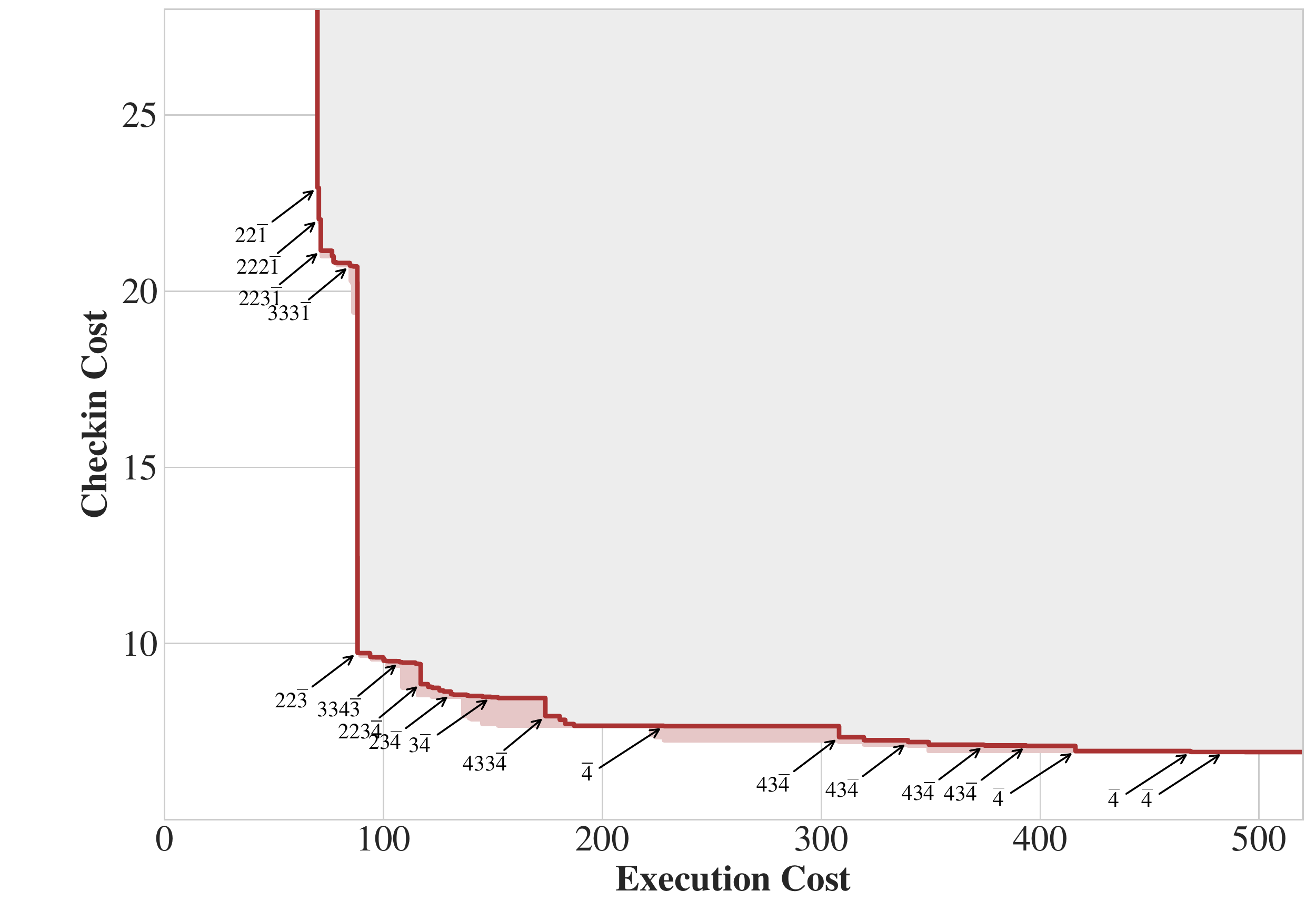}
        \caption{Up to length 4}
    \end{subfigure}
    \vspace{-2ex}
    
    \bigskip
    \begin{subfigure}{.45\linewidth}
        \centering
        \includegraphics[width=\textwidth]{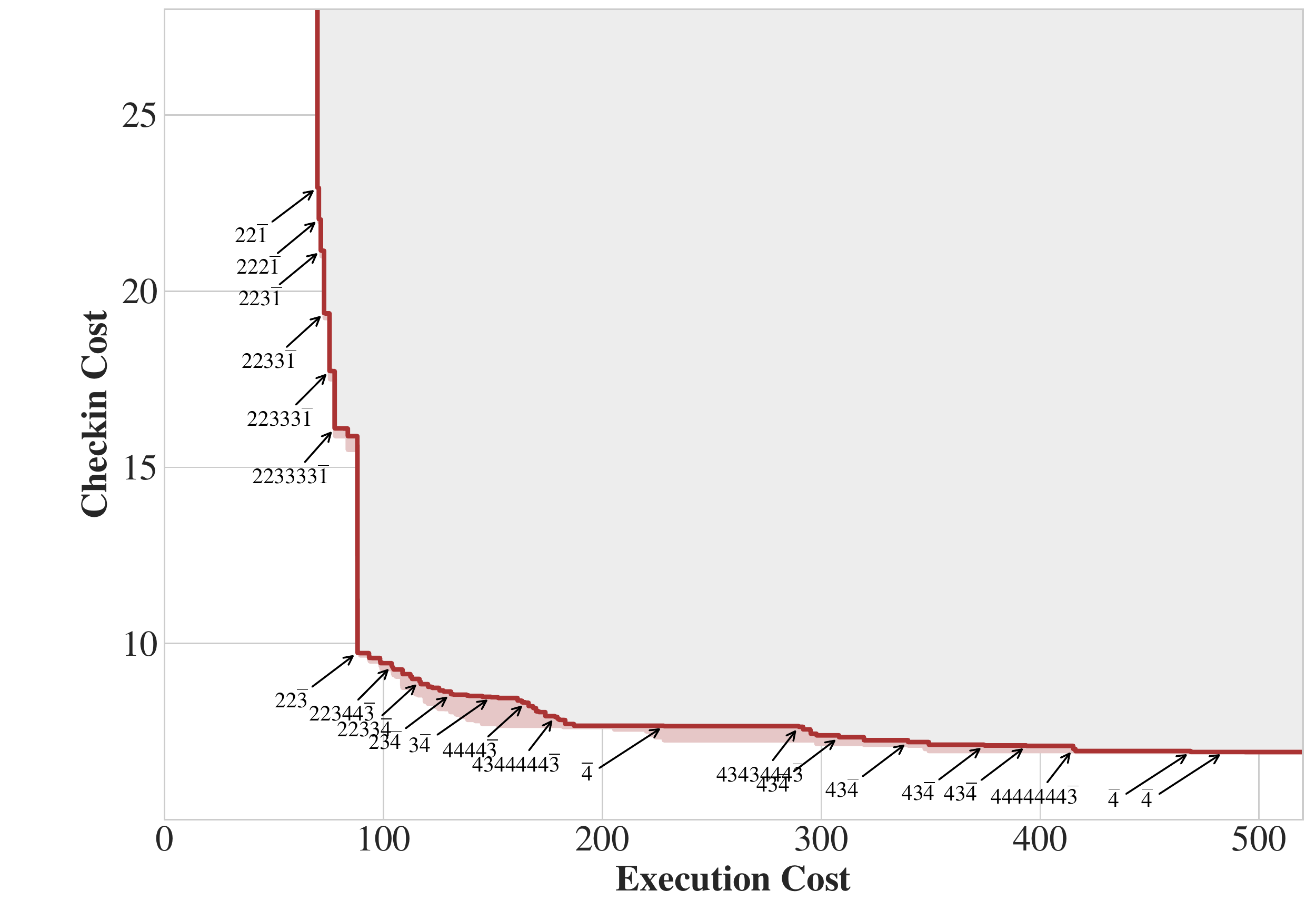}
        \caption{Up to length 8}
    \end{subfigure}
        \hfill
    \begin{subfigure}{.45\linewidth}
        \includegraphics[width=\textwidth]{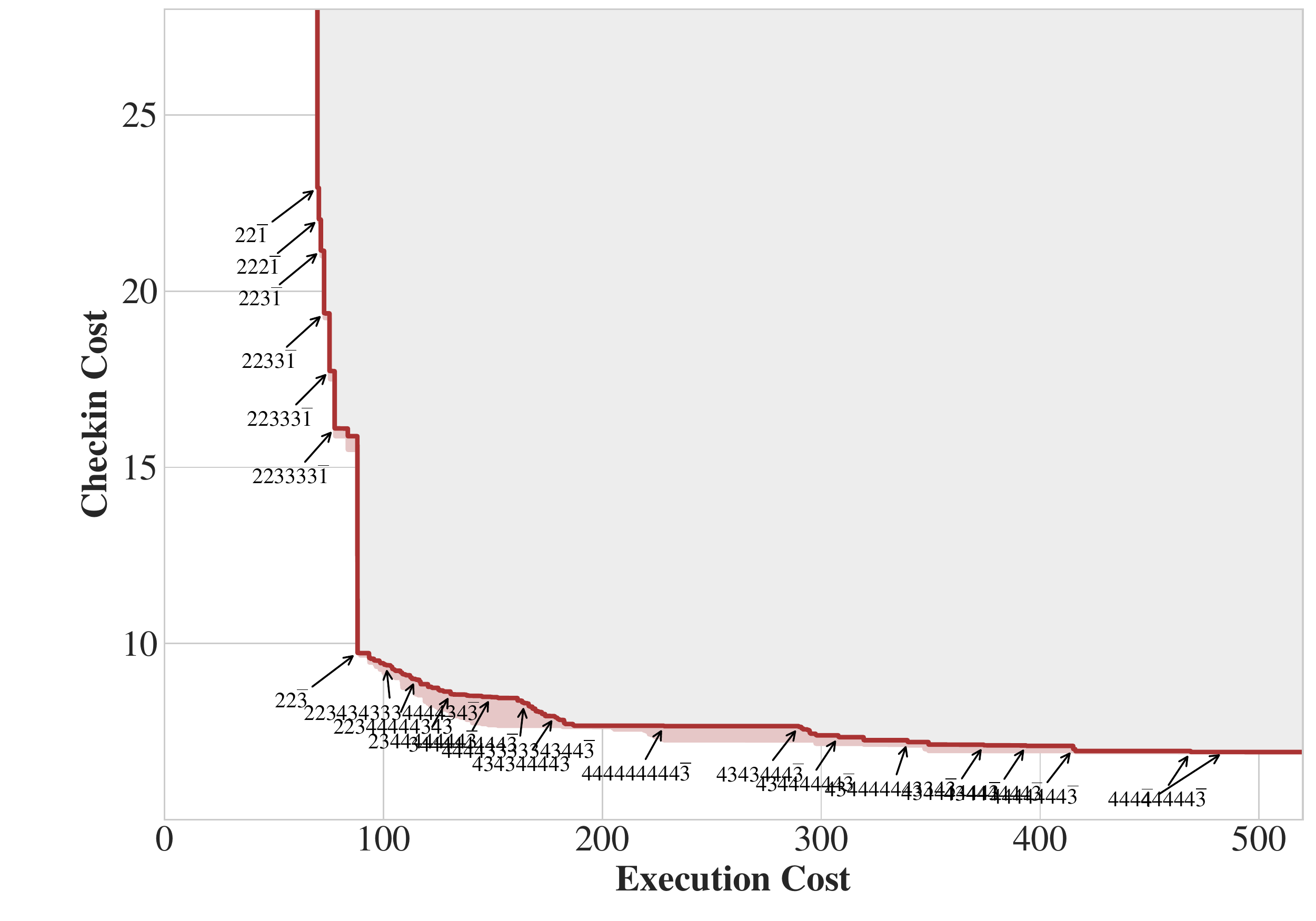}
        \caption{Up to length 16}
    \end{subfigure}

    \caption{Progression of the candidate Pareto front (red) as schedules get longer, up to length 16. Filtering using initial distribution with no margin tolerance and 10 alphas (0.1, 0.2, \dots, 0.9).\label{fig:pareto_c32_lengths}}
\end{figure}

In Figure~\ref{fig:pareto_c32_lengths} we examine an even larger problem as we extend schedules to length 16. Large changes in the front are made at the start, but as schedules get longer, fewer non-dominated schedules are discovered. 
As schedule lengths surpass the size of the environment, longer schedules become beneficial only for the less and less likely cases of not reaching the goal earlier.
In the figure we see that the longer non-dominated schedules become focused in a very small area. The filtering takes full advantage of this sparsity and diminishing returns. 
Otherwise, computing the entire roster of $4^{16}$ schedules would be quite impractical.

\section{Conclusion}
In tackling the problem of planning for a robot which relies on intermittent state observations from an external source, we examined methods for producing a pre-defined observation schedule suitable to both parties. 
The complication of execution and check-in costs being impacted by both the schedule and the policy to execute led us to formulate an approximate Pareto front, in which schedules are regions bounded by sub-fronts that can be improved through the use of alpha values.
We introduced a dynamic programming algorithm that constructs schedules via accumulation
through the use of a schedule extension procedure, computing incremental changes to value functions and policy evaluations. Further, we also proposed a filtering approach that prunes the working set to curb exponential growth. Our results demonstrated that this filtering scheme significantly reduces computation times for only negligible reductions in quality. Coupled with alpha values that tighten the bounds for more effective filtering and an overall performance gain for longer schedules, our algorithm allows for the computation of problem sizes that would not be feasible without filtering. 

\bibliographystyle{IEEEtranS}
\bibliography{refs}

\begin{thebibliography}{10}
\providecommand{\url}[1]{#1}
\csname url@samestyle\endcsname
\providecommand{\newblock}{\relax}
\providecommand{\bibinfo}[2]{#2}
\providecommand{\BIBentrySTDinterwordspacing}{\spaceskip=0pt\relax}
\providecommand{\BIBentryALTinterwordstretchfactor}{4}
\providecommand{\BIBentryALTinterwordspacing}{\spaceskip=\fontdimen2\font plus
\BIBentryALTinterwordstretchfactor\fontdimen3\font minus
  \fontdimen4\font\relax}
\providecommand{\BIBforeignlanguage}[2]{{%
\expandafter\ifx\csname l@#1\endcsname\relax
\typeout{** WARNING: IEEEtranS.bst: No hyphenation pattern has been}%
\typeout{** loaded for the language `#1'. Using the pattern for}%
\typeout{** the default language instead.}%
\else
\language=\csname l@#1\endcsname
\fi
#2}}
\providecommand{\BIBdecl}{\relax}
\BIBdecl

\bibitem{aloimonos93active}
Y.~Aloimonos, \emph{Active perception}.\hskip 1em plus 0.5em minus 0.4em\relax
  Mahwah, N.J.: LEA, Inc, 1993.

\bibitem{bajcsy1988active}
R.~Bajcsy, ``Active perception,'' \emph{Proceedings of the IEEE}, vol.~76,
  no.~8, pp. 966--1005, 1988.

\bibitem{bajcsy2018revisiting}
R.~Bajcsy, Y.~Aloimonos, and J.~K. Tsotsos, ``Revisiting active perception,''
  \emph{Autonomous Robots}, vol.~42, no.~2, pp. 177--196, 2018.

\bibitem{batalin2004mobile}
M.~A. Batalin, G.~S. Sukhatme, and M.~Hattig, ``{Mobile robot navigation using
  a sensor network},'' in \emph{IEEE International Conference on Robotics and
  Automation (ICRA)}, Apr. 2004, pp. 636--641.

\bibitem{bertsekas19reinforcement}
D.~P. Bertsekas, \emph{{Reinforcement Learning and Optimal Control}}.\hskip 1em
  plus 0.5em minus 0.4em\relax Belmont, M.A., U.S.A: Athena Scientific, 2019.

\bibitem{lavalle06planning}
S.~M. LaValle, \emph{{Planning Algorithms}}.\hskip 1em plus 0.5em minus
  0.4em\relax Cambridge, U.K.: Cambridge University Press, 2006, available at
  http://planning.cs.uiuc.edu/.

\bibitem{puterman2014markov}
M.~L. Puterman, \emph{Markov decision processes: discrete stochastic dynamic
  programming}.\hskip 1em plus 0.5em minus 0.4em\relax John Wiley \& Sons,
  2014.

\bibitem{rossi22periodic}
F.~Rossi and D.~A. Shell, ``Planning under periodic observations: bounds and
  bounding-based solutions,'' in \emph{IEEE/RSJ International Conference on
  Intelligent Robots and Systems (IROS)}, Oct. 2022.

\bibitem{roth2006communicate}
M.~Roth, R.~Simmons, and M.~Veloso, ``{What to communicate? Execution-time
  decision in multi-agent POMDPs},'' in \emph{Distributed Autonomous Robotic
  Systems (DARS) 7}.\hskip 1em plus 0.5em minus 0.4em\relax Springer, 2006, pp.
  177--186.

\bibitem{unhelkar2016contact}
V.~Unhelkar and J.~Shah, ``{Contact: Deciding to communicate during
  time-critical collaborative tasks in unknown, deterministic domains},'' in
  \emph{Proceedings of the AAAI Conference on Artificial Intelligence},
  vol.~30, no.~1, 2016.

\end{thebibliography}

\end{document}